\documentclass{article}

\PassOptionsToPackage{square,numbers}{natbib}

\usepackage[preprint]{neurips_2024}

\usepackage[utf8]{inputenc} %
\usepackage[T1]{fontenc}    %
\usepackage{hyperref}       %
\usepackage{url}            %
\usepackage{xurl}           %
\usepackage{booktabs}       %
\usepackage{tabularray}
\UseTblrLibrary{booktabs}
\usepackage[titles]{tocloft}
\usepackage{amsfonts}       %
\usepackage{amsmath}
\usepackage{amssymb}
\usepackage{nicefrac}       %
\usepackage{microtype}      %
\usepackage{geometry}
\usepackage{multirow}
\usepackage{fancyhdr}
\usepackage{graphicx}
\usepackage[table,xcdraw]{xcolor}
\usepackage{colortbl}
\usepackage{listings}
\usepackage{subcaption}
\usepackage{adjustbox}
\usepackage{caption}
\usepackage{wrapfig}
\usepackage[framemethod=TikZ]{mdframed}
\usepackage{multirow, makecell}
\usepackage{multicol}
\usepackage{svg}
\usepackage{numerica}
\hypersetup{
    colorlinks=true,
    linkcolor=darkgray,
    citecolor=gray,
    citebordercolor=0 0 0,
    urlcolor=darkgray
}

\newcommand{\ci}[1]{
{\scriptsize$\pm#1$}
}

\title{The Amazon Nova Family of Models:\\Technical Report and Model Card}

\author{%
Amazon Artificial General Intelligence
}

\begin{document}

\maketitle

\addtolength{\headwidth}{0.5in}
\chead{The Amazon Nova Family of Models}
\lhead{}
\rhead{}

\begin{figure}[h!]
\centering
\begin{subfigure}[c]{.45\linewidth}
    \centering
    \includegraphics[width=\textwidth]{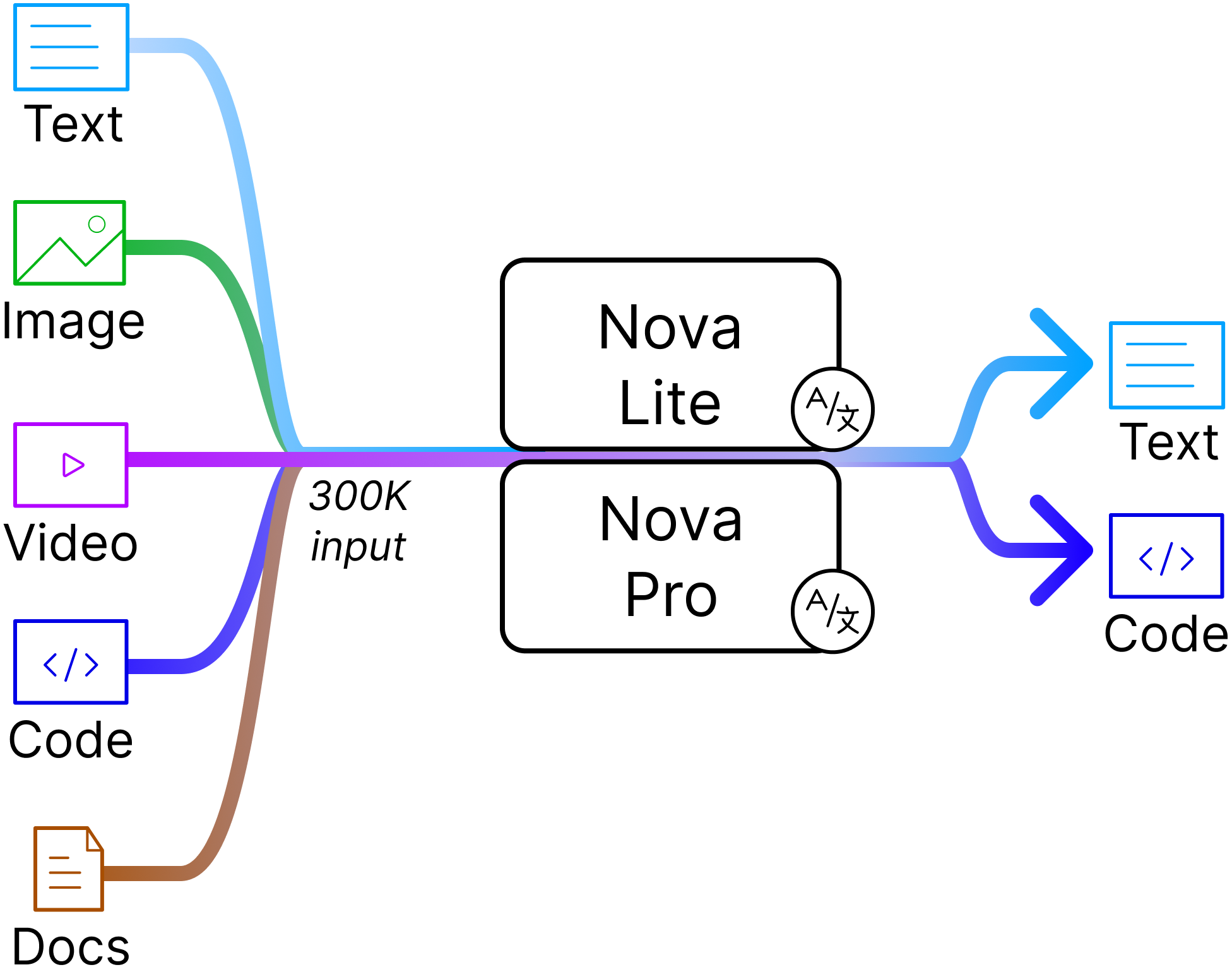}
\end{subfigure}
\hfill
\begin{subfigure}[c]{.45\linewidth}
    \centering
    \includegraphics[width=\textwidth]{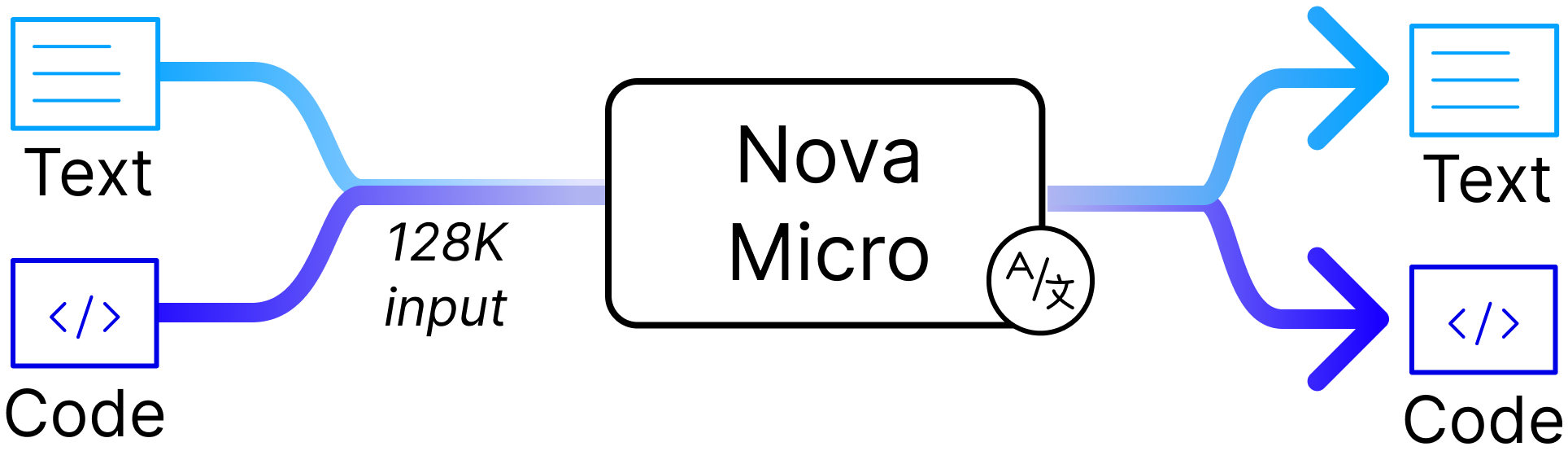}
\end{subfigure}
\par\medskip
\begin{subfigure}[c]{.45\linewidth}
    \centering
    \includegraphics[width=\textwidth]{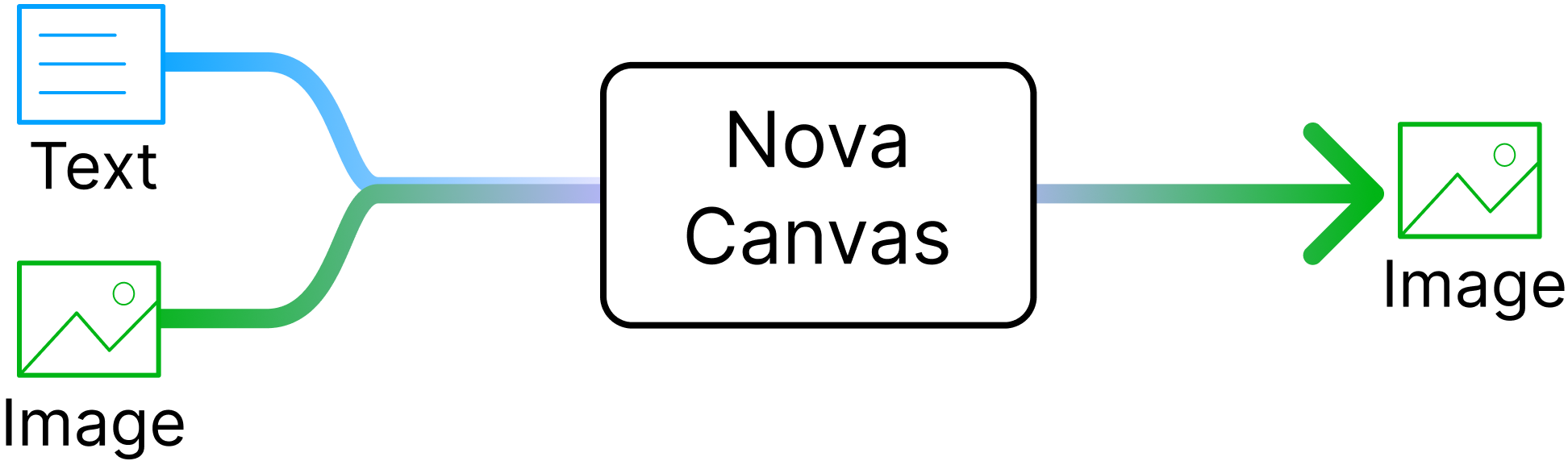}
\end{subfigure}
\hfill
\begin{subfigure}[c]{.45\linewidth}
    \centering
    \includegraphics[width=\textwidth]{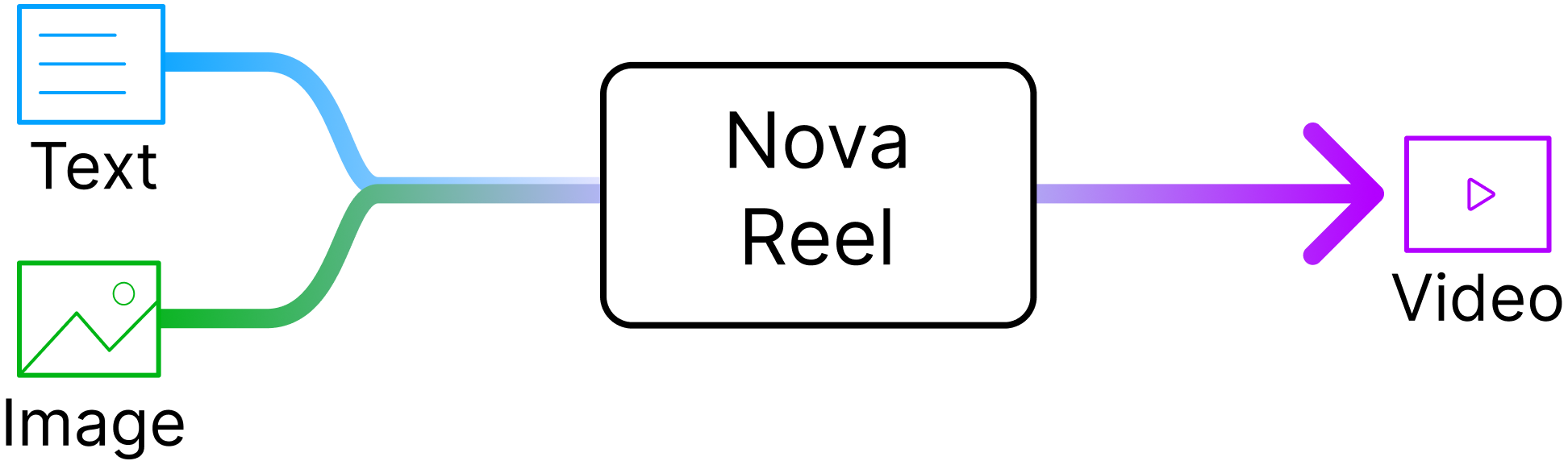}
\end{subfigure}
\par\medskip
\caption{The Amazon Nova family of models}
\end{figure}

\vspace{1cm}

\begin{abstract}
We present Amazon Nova, a new generation of state-of-the-art foundation models that deliver frontier intelligence and industry-leading price performance. 
Amazon Nova Pro is a highly-capable multimodal model with the best combination of accuracy, speed, and cost for a wide range of tasks.
Amazon Nova Lite is a low-cost multimodal model that is lightning fast for processing images, video, documents and text. 
Amazon Nova Micro is a text-only model that delivers our lowest-latency responses at very low cost. 
Amazon Nova Canvas is an image generation model that creates professional grade images with rich customization controls. 
Amazon Nova Reel is a video generation model offering high-quality outputs, customization, and motion control.
Our models were built responsibly and with a commitment to customer trust, security, and reliability.
We report benchmarking results for core capabilities, agentic performance, long context, functional adaptation, runtime performance, and human evaluation.
\end{abstract}

\newpage
\tableofcontents
\newpage

\section{Introduction}

This document introduces Amazon Nova, a new generation of state-of-the-art foundation models that deliver frontier intelligence and industry-leading price performance.

\subsection{Amazon Nova Pro, Lite, and Micro}

Key capabilities of Amazon Nova Pro, Lite, and Micro include:

\begin{itemize}
\item {\em Frontier intelligence:} Amazon Nova models possess frontier intelligence, enabling them to understand and process complex language tasks with state-of-the-art accuracy. Amazon Nova Micro sets new standards in its intelligence tier in several text benchmarks such as Language Understanding (MMLU), Deep Reasoning (GPQA), Mathematics (MATH), and Multi-step Reasoning (Big-Bench Hard). Our multimodal models, Amazon Nova Pro and Lite, take text, images, documents, and video as input and generate text as output. These models set standards in several benchmarks such as Video Captioning (VATEX), Visual QA (TextVQA), Function Calling (BFCL), and multimodal agentic benchmarks (GroundUI-1K, VisualWebBench, Mind2Web) in their respective intelligence tiers. These models are the first to offer video understanding capabilities on Amazon Bedrock, enabling deeper insights from multimedia content. 
\item {\em Speed:} Amazon Nova has been designed for fast inference, with Amazon Micro, Lite, and Pro each being one of the fastest models in their respective intelligence tiers.
\item {\em Agentic Workflows:} Amazon Nova Pro, Lite, and Micro can power AI agents capable of breaking down and executing multi-step tasks. These models are integrated with Bedrock Knowledge Bases and they excel at retrieval-augmented generation (RAG) to ensure the best accuracy by grounding their responses to the developer's data.
\item {\em Customizability:} Developers can fine-tune these models with multimodal data (Pro and Lite) or text data (Pro, Lite, and Micro), providing the flexibility to achieve desired accuracy, latency, and cost. Developers can also run self-service Custom Fine-Tuning (CFT) and distillation of larger models to smaller ones via Bedrock APIs. 
\item {\em Price-Performance:} Each model was optimized to deliver exceptional price-performance value, offering state-of-the-art performance on key benchmarks at low cost.
\end{itemize}

Amazon {Nova Pro}, {Lite}, and {Micro} are based on the Transformer architecture \cite{vaswani2023attentionneed}.
Each model went through a series of training processes that began with pretraining using a mixture of large amounts of multilingual and multimodal data.
Our models were trained on data from a variety of sources, including licensed data, proprietary data, open source datasets, and publicly available data where appropriate.
We curated data from over 200 languages, with particular emphasis on Arabic, Dutch, English, French, German, Hebrew, Hindi, Italian, Japanese, Korean, Portuguese, Russian, Simplified Chinese, Spanish, and Turkish.
After pretraining, models iteratively went through a series of fine-tuning stages, including Supervised Fine-Tuning (SFT) on instruction-demonstration pairs (including multimodal ones) and reward model (RM) training from human preference data \cite{Ouyang22TL}.
Finally, the models learned from human preferences via methods like Direct Preference Optimization (DPO) \cite{Rafailov23DP} and Proximal Policy Optimization (PPO) \cite{Schulman17PP} to ensure that the final models are aligned with human preferences in both quality and responsibility.

\subsection{Amazon Nova Canvas and Reel}

Amazon Nova Canvas and Amazon Nova Reel are designed to create realistic multimodal content, including images and videos, for a wide range of applications such as advertising, marketing, and entertainment.

Amazon Nova Canvas offers the following functionalities, with more details provided in Appendix \ref{appendix_canvas}: 
\begin{itemize}
\item {\em Text-to-image generation:} Amazon Nova Canvas can generate images with various resolutions (from 512 up to 2K horizontal resolution) and aspect ratios (any aspect ratio between 1:4 and 4:1 with a maximum of 4.2M pixels). Customers can provide reference images to guide the model to generate outputs in a specific style or color palette, or to generate variations of an image.
\item {\em Image editing:} Amazon Nova Canvas allows precise image editing operations like inpainting and outpainting through natural language mask prompts. These mask prompts describe the specific area of the input image that needs to be repainted. The user can also easily change a background with the background removal feature, leaving the subject of the image unchanged. 
\end{itemize}

\begin{samepage}
Amazon Nova Reel offers the following functionalities:
\begin{itemize}
\item {\em Generate videos from a text prompt:} Amazon Nova Reel can generate high-quality videos of 6-second duration (720p resolution at 24 frames per second) from a text prompt.
\item {\em Generate videos from a reference image and a prompt:} Amazon Nova Reel brings images to motion and generates videos that are guided by the input image and a text prompt.
\item {\em Camera motion control using a text prompt:} With camera motion control in Amazon Nova Reel, the user can guide camera motion with text prompts like ``zoom'' and ``dolly forward'' to get the exact visual needed for each video. Amazon Nova Reel supports more than 20 camera motions. For more details, please refer to our prompting guide\footnote{\url{https://docs.aws.amazon.com/nova/latest/userguide}}.
\end{itemize}
\end{samepage}

Amazon {Nova Canvas} and {Reel} are latent diffusion models \cite{dit} where a Variational AutoEncoder (VAE) \cite{kingma2013auto} maps the image or video frames to latent variables on which the diffusion process happens.
A text encoder tokenizes input text prompts into tokens which are then passed to the diffusion model as a conditioning signal. At inference time, a latent variable is initialized with random noise sampled from a Gaussian distribution, which is then denoised by the trained diffusion model iteratively into a clean latent variable. The clean latent variable is decoded back to images or video frames by the decoder of the VAE. Both models underwent a two-phased approach of pretraining and fine-tuning. Pretraining data were sourced from a variety of sources, including licensed data, proprietary data, open source datasets, and publicly available data where appropriate. Our highly scalable data filtering, deduplication, and enrichment pipelines were based on AWS EMR~\cite{aws_emr} and AWS Batch~\cite{aws_batch}, as well as other AWS services.

\clearpage
\section{Amazon Nova Pro, Lite, and Micro Evaluations}
In this section, we report benchmarking results for Amazon Nova models and for select publicly-available models, including by citing existing public results and by measuring their performance.\footnote{\label{api_measurements}Results measured internally by Amazon for evaluation purposes after Amazon Nova models completed training using (i) the Bedrock API for Claude and Meta models or (ii) the OpenAI API or Gemini API, as applicable.} 
In cases for which the result is a simple average of binary scores, we assume a Gaussian distribution for the sample and approximate the 95\% confidence interval as:
\begin{equation}
    CI(S) = 1.96 \times \sqrt{\frac{S\times(1-S)}{N}}
\end{equation}

where $CI$ is the 95\% confidence interval, $S$ is the measured score for the benchmark, and $N$ is the sample size \cite{madaan2024,llama31}.

\subsection{Core capability public benchmarks}

We evaluate Amazon Nova models on a suite of automated public benchmarks to assess core capabilities, including for both text-only (Section \ref{ssec:core_text}) and multimodal (Section \ref{sec:evaluations_multimodal_intelligence}) use cases. 

\subsubsection{Core capability text benchmarks and results}
\label{ssec:core_text}

We evaluate select core capabilities of Amazon Nova models on a variety of public text-only benchmarks, spanning general knowledge, reasoning, language understanding, multilinguality, and instruction following. 

\begin{table}[p]
\centering
\begin{tblr}{
    colspec={lX[-1,c]X[1,c]X[1,c]X[1,c]X[1,c]X[1,c]X[1,c]X[1,c]X[1,c]},
    rows={m},
    row{1}={font=\bfseries\scriptsize, valign=m},
    column{2}={font=\footnotesize},
    row{2}={font=\scriptsize\itshape, valign=m},
    row{6}={font=\scriptsize},
    row{9}={font=\scriptsize},
    row{13}={font=\scriptsize},
    row{16}={font=\scriptsize},
    row{20}={font=\scriptsize},
}
\toprule
   &     &   MMLU   &   ARC-C   &   DROP   &   GPQA   &   MATH   &   GSM8k   &   IFEval   &   BBH \\   
   &   {tok/\\sec}   &   accuracy   &   accuracy   &   F1-score   &   accuracy   &   accuracy   &   accuracy   &   instruction-level loose accuracy   &   accuracy \\
\midrule
Nova Pro     & 100 & 85.9 &  94.8\ci{1.3}  &  85.4\ci{0.7}  &  46.9\ci{4.6}  &  76.6\ci{1.2}  & 94.8\ci{1.2} & 92.1\ci{1.8} &   86.9 \\
Nova Lite    & 157 & 80.5 &  92.4\ci{1.5}  &  80.2\ci{0.8}  &  42.0\ci{4.6}  &  73.3\ci{1.2}  & 94.5\ci{1.2} & 89.7\ci{2.1} &   82.4 \\
Nova Micro   & 210 & 77.6 &  90.2\ci{1.7}  &  79.3\ci{0.8}  &  40.0\ci{4.5}  &  69.3\ci{1.3}  & 92.3\ci{1.4} & 87.2\ci{2.3} &   79.5 \\
   &     &   0-shot CoT   &   0-shot   &   6-shot CoT   &   0-shot CoT   &   0-shot CoT   &   0-shot CoT   &   0-shot   &   {3-shot\\CoT}  \\ 
\midrule
Claude 3.5 Sonnet (Oct)    & 57 & 89.3 &  96.3\TblrNote{$\scriptstyle \textit{M}$}\ci{1.1}  &  88.3\ci{0.6}   &  58.0\TblrNote{$\scriptstyle \textit{M}$}\ci{4.6}  &  78.3\ci{1.1}  &   96.5\TblrNote{$\scriptstyle \textit{M}$}\ci{1.0} & 90.2$\scriptstyle ^\textit{*}$\ci{2.0} &   93.2 \\
Claude 3.5 Haiku   & 64 & 80.3 &  90.9\TblrNote{$\scriptstyle \textit{M}$}\ci{1.6}  &   83.1\ci{0.8}  &  37.5\TblrNote{$\scriptstyle \textit{M}$}\ci{4.5}  &  69.4\ci{1.3}  &   93.8\TblrNote{$\scriptstyle \textit{M}$}\ci{1.3}   & 85.9$\scriptstyle ^\textit{*}$\ci{2.4} &   86.6 \\
   &     &   0-shot CoT   &   25-shot  &   3-shot   &   0-shot CoT   &   0-shot CoT   &   0-shot CoT   &   0-shot   &   {3-shot\\CoT}  \\ 
\midrule
Gemini 1.5 Pro (002)   & 58 & 85.9 &  95.4\TblrNote{$\scriptstyle \textit{M}$}\ci{1.2}  &   74.9\ci{0.9}   &  55.1\TblrNote{$\scriptstyle \textit{M}$}\ci{4.6}  &  86.5\ci{0.9}  & 90.8\ci{1.6} &   91.7\TblrNote{$\scriptstyle \textit{M}$}\ci{1.9}   &   89.2 \\
Gemini 1.5 Flash (002)   & 190 & 78.9 &  94.3\TblrNote{$\scriptstyle \textit{M}$}\ci{1.3}  &   78.4\ci{0.8}   &  45.1\TblrNote{$\scriptstyle \textit{M}$}\ci{4.6}  &  77.9\ci{1.2}  & 86.2\ci{1.9} &   91.6\TblrNote{$\scriptstyle \textit{M}$}\ci{1.9}   &   85.5 \\
Gemini 1.5 Flash 8B (001)   &   283   & 68.1 &  88.7\TblrNote{$\scriptstyle \textit{M}$}\ci{1.8}  &   68.1\TblrNote{$\scriptstyle \textit{M}$}\ci{0.9}   &  33.5\TblrNote{$\scriptstyle \textit{M}$}\ci{4.4}  &  58.7\ci{1.4}  & 84.5\TblrNote{$\scriptstyle \textit{M}$}\ci{2.0} &   86.1\TblrNote{$\scriptstyle \textit{M}$}\ci{2.3}   &   69.5 \\
   &     &   5-shot   &   25-shot   &   3-shot   &   0-shot   &   4-shot   &   11-shot   &   0-shot   &   3-shot  \\ 
\midrule
GPT-4o   & 163 & 88.7 &  96.2\TblrNote{$\scriptstyle \textit{M}$}\ci{1.1}  &  83.4\ci{0.7}  &  48.4\TblrNote{$\scriptstyle \textit{M}$}\ci{4.6}  &  76.6\ci{1.2}  &   92.6\TblrNote{$\scriptstyle \textit{M}$}\ci{1.4}   &   89.8\TblrNote{$\scriptstyle \textit{M}$}\ci{2.1}   &   83.0\TblrNote{$\scriptstyle \textit{M}$} \\
GPT-4o Mini   & 113 & 82.0 &  92.3\TblrNote{$\scriptstyle \textit{M}$}\ci{1.5}  &  79.7\ci{0.8}  &  41.7\TblrNote{$\scriptstyle \textit{M}$}\ci{4.6}  &  70.2\ci{1.3}  &   86.4\TblrNote{$\scriptstyle \textit{M}$}\ci{1.8} &  87.4\TblrNote{$\scriptstyle \textit{M}$}\ci{2.3}   &   81.0\TblrNote{$\scriptstyle \textit{M}$} \\
   &     &   0-shot   &   25-shot   &   3-shot   &   0-shot   &   0-shot CoT   &   0-shot CoT   &   0-shot   &   3-shot  \\ 
\midrule
Llama 3.2 90B   & 40 & 86.0 &  94.8\ci{1.3}  &   -   &   46.7\ci{4.6}   &  68.0\ci{1.3}  & 95.1\ci{1.2} &   90.9\TblrNote{$\scriptstyle \textit{M}$}\ci{2.0}   &   - \\
Llama 3.2 11B  &  124  & 73.0  &  83.4\ci{2.1}  &  -  &  32.8\ci{4.3}  &  51.9\ci{1.4}  &  84.5\ci{2.0}  &  85.0\TblrNote{$\scriptstyle \textit{M}$}\ci{2.4}  & - \\
Llama 3.1 8B   & 157 & 73.0 &  83.4\ci{2.1}  &   -   &   30.4\ci{4.3}   &  51.9\ci{1.4}  & 84.5\ci{2.0} &   85.0\TblrNote{$\scriptstyle \textit{M}$}\ci{2.4}  &   -   \\
   &     &   0-shot CoT   &   25-shot   &   -   &   0-shot   &   0-shot CoT   &   8-shot CoT   &   -   &   -  \\ 
\bottomrule
\end{tblr}
\caption[lorem]{Quantitative results on core capability benchmarks (MMLU~\cite{mmlu}, ARC-C~\cite{allenai:arc}, DROP~\cite{Dua2019DROP}, GPQA~\cite{rein2023gpqagraduatelevelgoogleproofqa}, MATH~\cite{hendrycksmath2021}), GSM8K~\cite{cobbe2021gsm8k}, IFEval~\cite{zhou2023instructionfollowingevaluationlargelanguage} and BigBench-Hard (BBH)~\cite{suzgun2022challenging}). Unless otherwise noted, all reference numbers are taken from the original technical reports and websites for Claude models~\cite{claude_addendum, claude3modelfamily}, GPT4 models~\cite{o1mini, GPT4evals}, Llama models~\cite{llama31} and Gemini models~\cite{gemini_1_5}. Results marked with $\scriptstyle \textit{M}$ were measured by us\textsuperscript{\ref{api_measurements}}. Claude numbers for IFEval (taken from~\cite{claude_addendum}) are marked with an asterisk ($*$), as the scoring methodology is unspecified in the report. Token generation speed in tokens per second (tok/sec), the inverse of per-token generation latency, is reproduced from Section~\ref{sec:eval_runtime}.}
\label{tab:evaluations_general_intelligence}

\end{table}

The following list briefly describes our selected text-only benchmarks. The prompts used for evaluation of each benchmark are summarized in Appendix~\ref{sec:appendix_prompts_text_general}.
\begin{itemize}
    \item {MMLU~\cite{mmlu}}: Massive Multitask Language Understanding (MMLU) is a multiple-choice question answering benchmark that covers 57 subject areas across STEM, humanities, and social sciences. Subjects include law, physics, mathematics, computer science, history, and more. The difficulty levels vary from elementary level to advanced professional level, focusing on both world knowledge and problem solving abilities. We use 0-shot Chain-of-Thought (CoT) \cite{cot} for prompting and report the macro average exact match accuracy across all subjects. 
    \item {ARC-C~\cite{allenai:arc}}: The AI2’s Reasoning Challenge (ARC) is a multiple-choice question-answering dataset, which contains science questions from grade 3 to grade 9 exams. We use 0-shot CoT for prompting and report exact match accuracy.
    \item {DROP~\cite{Dua2019DROP}}: Discrete Reasoning Over Paragraphs (DROP) is a crowdsourced reading comprehension dataset that requires reasoning and operating over multiple input positions from the reference text. We use 0-shot CoT for prompting and report f1 score. 
    \item {GPQA~\cite{rein2023gpqagraduatelevelgoogleproofqa}:} Graduate-level Google-Proof Question and Answering (GPQA) is a challenging and high-quality multiple-choice question answering benchmark written by domain experts who have or are pursuing PhDs in biology, physics, and chemistry. We use 0-shot CoT for prompting and report exact match accuracy on the main set. 
    \item {MATH~\cite{hendrycksmath2021}:} MATH is a mathematics problem solving benchmark, consisting of problems from mathematics competitions including the American Mathematics Competitions (AMC 10 and AMC 12), the American Invitational Mathematics Examination (AIME) and more. We use 0-shot CoT for prompting and report the exact match accuracy on the MATH5k set.
    \item {GSM8K~\cite{cobbe2021gsm8k}:} Grade School Math 8K (GSM8K) is a math benchmark consisting of 8,500 high-quality and diverse grade school math problems. The benchmark tests basic mathematical problem solving capabilities, requiring multi-step reasoning. We use 0-shot CoT for prompting and report the exact match accuracy on the test set containing 1,319 samples. 
    \item {IFEval ~\cite{zhou2023instructionfollowingevaluationlargelanguage}:} 
    IFeval is an instruction-following benchmark, which evaluates a model’s capability of following ``verifiable instructions'' such as ``mention the keyword of AI at least 3 times''. The dataset contains 25 types of verifiable instructions and in total 541 prompts, where each prompt contains one or more verifiable instructions in natural language. We report the instruction-level accuracy under loose constraints. 
    \item {BBH~\cite{suzgun2022challenging}:} Big Bench Hard (BBH) is a diverse benchmark consisting of an aggregate of 23 diverse subjects that cover algorithmic and NLP tasks ranging from casual logic tasks to word sorting and movie recommendations. The tasks are both multiple choice and open generation tasks. We report the macro average exact match accuracy across the subjects. 
\end{itemize}

Table~\ref{tab:evaluations_general_intelligence} summarizes the quantitative results of Nova models and select public models on the aforementioned benchmarks for core capabilities. When available, we reference the highest publicly-reported numbers for each benchmark from the official technical reports and websites for Claude, Gemini, OpenAI and Llama family of models. Amazon Nova Pro, Lite, and Micro demonstrate strong performance across all benchmarks, showcasing their advanced core intelligence, particularly Amazon Nova Micro and Lite on math, reasoning, and instruction following benchmarks. 

We also evaluate the translation capabilities of Nova models. Flores200~\cite{nllb2022, flores101, flores_remaining}, or simply Flores, is a machine translation benchmark consisting of translations from 842 distinct web articles, which tests the translation capabilities between English and non-English languages. Sentences are 21 words long on average. We use a 0-shot setup and report the macro average of two metrics, spBleu and COMET22 score~\cite{comet22} across a set of languages (Arabic, German, Spanish, French, Hindi, Italian, Japanese, Korean, Portuguese, Hebrew, Turkish, Simplified Chinese, Russian, Dutch) for translation from and into English. The prompts used for evaluation are summarized in Appendix~\ref{sec:appendix_prompts_text_general}.
Table~\ref{tab:evaluation_multilingual} summarizes our quantitative results on Flores, demonstrating strong multilingual performance on translation for Amazon Nova Pro, Lite, and Micro.

\begin{table}[ht]
\centering
\begin{tblr}{
  width=0.9\textwidth,
  colspec={lX[-1,c]X[1,c]X[1,c]X[1,c]X[1,c]},
  rows={m},
  column{2} = {font=\footnotesize},
  row{3}={font=\footnotesize\itshape},
}
\toprule
& & \SetCell[c=4]{c}{\textbf{FLORES (0-shot)}} & & &  \\
& & \SetCell[c=2]{c}{en $\rightarrow$ Set1} &
& \SetCell[c=2]{c}{Set1 $\rightarrow$ en} & 
\\
& tok/sec & spBleu ($\uparrow$) & COMET22 ($\uparrow$) & spBleu ($\uparrow$) & COMET22 ($\uparrow$) \\
\midrule
Nova Pro & 100   & 43.4 & 89.1 & 44.4 & 89.0  \\
Nova Lite  & 157 & 41.5 & 88.8 & 43.1 & 88.8   \\
Nova Micro & 210 & 40.2 & 88.5 & 42.6 & 88.7  \\
\midrule
Claude 3.5 Sonnet (Oct) & 57   & 42.5\TblrNote{$\scriptstyle \textit{M}$} & 89.4\TblrNote{$\scriptstyle \textit{M}$} & 43.5\TblrNote{$\scriptstyle \textit{M}$} & 89.1\TblrNote{$\scriptstyle \textit{M}$}   \\
Claude 3.5 Haiku & 64  & 40.0\TblrNote{$\scriptstyle \textit{M}$} & 88.5\TblrNote{$\scriptstyle \textit{M}$} & 40.2\TblrNote{$\scriptstyle \textit{M}$} & 88.3\TblrNote{$\scriptstyle \textit{M}$} \\
\midrule
Gemini 1.5 Pro (002) & 57 & 43.0\TblrNote{$\scriptstyle \textit{M}$$\scriptstyle \textit{*}$} & 89.1\TblrNote{$\scriptstyle \textit{M}$$\scriptstyle \textit{*}$} & 45.6\TblrNote{$\scriptstyle \textit{M}$$\scriptstyle \textit{*}$} & 89.1\TblrNote{$\scriptstyle \textit{M}$$\scriptstyle \textit{*}$} \\
Gemini 1.5 Flash (002) & 190 & 40.0\TblrNote{$\scriptstyle \textit{M}$$\scriptstyle \textit{*}$} & 88.5\TblrNote{$\scriptstyle \textit{M}$$\scriptstyle \textit{*}$} & 42.9\TblrNote{$\scriptstyle \textit{M}$$\scriptstyle \textit{*}$} & 88.8\TblrNote{$\scriptstyle \textit{M}$$\scriptstyle \textit{*}$}  \\
Gemini 1.5 Flash 8B (001) & 283 & 38.2\TblrNote{$\scriptstyle \textit{M}$$\scriptstyle \textit{*}$} & 88.0\TblrNote{$\scriptstyle \textit{M}$$\scriptstyle \textit{*}$} & 41.4\TblrNote{$\scriptstyle \textit{M}$$\scriptstyle \textit{*}$} & 88.5\TblrNote{$\scriptstyle \textit{M}$$\scriptstyle \textit{*}$} \\
\midrule 
GPT-4o  & 163 & 43.1\TblrNote{$\scriptstyle \textit{M}$$\scriptstyle \textit{*}$} & 89.2\TblrNote{$\scriptstyle \textit{M}$$\scriptstyle \textit{*}$} & 43.9\TblrNote{$\scriptstyle \textit{M}$$\scriptstyle \textit{*}$} & 89.0\TblrNote{$\scriptstyle \textit{M}$$\scriptstyle \textit{*}$}   \\
GPT-4o Mini  & 113 & 41.1\TblrNote{$\scriptstyle \textit{M}$$\scriptstyle \textit{*}$} & 88.7\TblrNote{$\scriptstyle \textit{M}$$\scriptstyle \textit{*}$} & 41.9\TblrNote{$\scriptstyle \textit{M}$$\scriptstyle \textit{*}$} & 88.7\TblrNote{$\scriptstyle \textit{M}$$\scriptstyle \textit{*}$}  \\
\midrule
Llama 3.2 90B  & 40 & 39.7\TblrNote{$\scriptstyle \textit{M}$} & 88.2\TblrNote{$\scriptstyle \textit{M}$} & 43.7\TblrNote{$\scriptstyle \textit{M}$} & 88.5\TblrNote{$\scriptstyle \textit{M}$}   \\
Llama 3.2 11B & 124 & 33.0\TblrNote{$\scriptstyle \textit{M}$} & 85.7\TblrNote{$\scriptstyle \textit{M}$} & 36.3\TblrNote{$\scriptstyle \textit{M}$} & 86.3\TblrNote{$\scriptstyle \textit{M}$}  \\
Llama 3.1 8B  & 157 & 32.7\TblrNote{$\scriptstyle \textit{M}$} & 85.5\TblrNote{$\scriptstyle \textit{M}$} & 36.5\TblrNote{$\scriptstyle \textit{M}$} & 86.5\TblrNote{$\scriptstyle \textit{M}$}  \\

\bottomrule
\end{tblr}
\caption[lorem]{Quantitative results on Flores200~\cite{flores101}, a machine translation benchmark. Set1 refers to \{de, es, fr, it, pt, ja, ar, hi, ru, nl, tr, he, ko, zh\}. %
Results marked with $\scriptstyle \textit{M}$ were measured by us.\textsuperscript{\ref{api_measurements}}. Results marked with an asterisk ($*$) were obtained using an alternate prompt which can be found in Appendix~\ref{sec:appendix_prompts_text_general} Token generation speed in tokens per second (tok/sec), the inverse of per-token generation latency, is reproduced from Section~\ref{sec:eval_runtime}.}
\label{tab:evaluation_multilingual}

\end{table}

\subsubsection{Core capability multimodal benchmarks and results}
\label{sec:evaluations_multimodal_intelligence}

\begin{table}[htp]
\centering
\footnotesize
\begin{tblr}{
    colspec={lX[-1,c]X[1,c]X[1,c]X[1,c]X[1,c]X[1,c]X[1,c]},
    rows={m},
    column{2}={font=\scriptsize},
    row{1}={font=\bfseries},
    row{3}={font=\itshape},
}
\toprule
& & MMMU (CoT) & {Chart\\QA\TblrNote{$\scriptstyle \textit{C}$}} & {Doc\\VQA} & {Text\\VQA} & VATEX & {Ego\\Schema} \\
& & val & test & test & val & test & test \\
& {tok/\\sec} & accuracy & relaxed accuracy & ANLS & weighted accuracy & CIDEr & accuracy \\
\midrule
Amazon Nova Pro  & 100 & 61.7\ci{3.2} & 89.2\ci{1.2} & 93.5 & 81.5 & 77.8 & 72.1\ci{5.4} \\
Amazon Nova Lite & 157 & 56.2\ci{3.2} & 86.8\ci{1.3} & 92.4 & 80.2 & 77.8 & 71.4\ci{5.4} \\
\midrule
Claude 3.5 Sonnet (Oct)  & 57 & 70.4\ci{3.0} & 90.8\ci{1.1} & 94.2 & 61.7\TblrNote{$\scriptstyle \textit{M}$} & - & - \\
Claude 3 Haiku & 64 & 50.2\ci{3.3} & 82.0\ci{1.5} & 88.8 & - & - & - \\
\midrule
Gemini 1.5 Pro (001)  & 58 & 65.9\ci{3.1}\TblrNote{$\scriptstyle \textit{E}$} & 87.2\ci{1.3} & 93.1\TblrNote{$\scriptstyle \textit{B}$} & 78.7 & 64.6\TblrNote{$\scriptstyle \textit{A}$} & 72.2\ci{5.4} \\
Gemini 1.5 Flash (001)  & 190 & 62.3\ci{3.2}\TblrNote{$\scriptstyle \textit{E}$} & 85.4\ci{1.4} & 89.9\TblrNote{$\scriptstyle \textit{B}$} & 78.7 & 57.1 & 65.7\ci{5.7} \\
Gemini 1.5 Flash 8B (001)  & 283 & 53.7\ci{3.3}\TblrNote{$\scriptstyle \textit{F}$} & 78.2\ci{1.6}\TblrNote{$\scriptstyle \textit{G}$} & 73.6 & 66.7 & 53.2\TblrNote{$\scriptstyle \textit{A}$} & - \\
\midrule
GPT-4o (May) & - & 69.1\ci{3.0} & 85.7\ci{1.4} & 92.8 & 77.2\TblrNote{$\scriptstyle \textit{D,M}$} & - & 72.2\ci{5.4} \\
GPT-4o Mini (Jul)  & 113 & 59.4\ci{3.2} & 79.2\ci{1.6}\TblrNote{$\scriptstyle \textit{M}$} & - & 70.3\TblrNote{$\scriptstyle \textit{M}$} & - & - \\
\midrule
Llama 3.2 90B  & 40 & 60.3\ci{3.2} & 85.5\ci{1.4} & 90.1 & 80.7\TblrNote{$\scriptstyle \textit{M}$} & - & - \\
Llama 3.2 11B  & 124 & 50.7\ci{3.3} & 83.4\ci{1.5} & 88.4 & 71.3\TblrNote{$\scriptstyle \textit{M}$} & - & - \\
\bottomrule
\end{tblr}
\caption{
Quantitative results on four image understanding benchmarks (MMMU~\cite{mmmu}, ChartQA~\cite{chartqa}, DocVQA~\cite{docvqa}, TextVQA~\cite{textvqa}) and 2 video understanding benchmarks (VATEX~\cite{vatex} and EgoSchema~\cite{egoschema}). Higher numbers are better for all benchmarks ($\uparrow$). Unless otherwise noted, all evaluations are 0-shot and reference numbers are taken from the original technical reports and websites for Claude models~\cite{claude3modelfamily,website_claude_sonnet}, GPT4 models~\cite{website_hello_gpt4o,website_gpt4o_mini}, Llama models~\cite{llama31,website_llama32_model_card_vision} and Gemini models~\cite{gemini_1_5,website_gemini_flash}.
Remarks:
($\scriptstyle \textit{A}$) 4-shot evaluation;
($\scriptstyle \textit{B}$) External Optical Character Recognition (OCR) was used;
($\scriptstyle \textit{C}$) All models except Amazon Nova use CoT;
($\scriptstyle \textit{D}$) GPT-4o (Nov);
($\scriptstyle \textit{E}$) Gemini 1.5 Flash/Pro (002) models;
($\scriptstyle \textit{F}$) Reported in \cite{website_gemini_flash};
($\scriptstyle \textit{G}$) Reported in \cite{agrawal2024pixtral12b};
($\scriptstyle \textit{M}$) Claude 3.5 Sonnet and Llama 3.2 results for TextVQA as well as GPT4o and GPT4o mini results on ChartQA, TextVQA and VATEX were measured by us.\textsuperscript{\ref{api_measurements}}
Token generation speed in tokens per second (tok/sec), the inverse of per-token generation latency, is reproduced from Section~\ref{sec:eval_runtime}.
}
\label{tab:evaluations_multimodal_intelligence}
\end{table}

In this section we evaluate the multimodal capabilities of Amazon Nova models on a diverse set of public benchmarks. Our selection of multimodal benchmarks aims to probe for various capabilities, including natural image understanding, document understanding with charts and graphs, text understanding, and temporal reasoning in videos. For all benchmarks, we follow the suggested metrics and choice of data split for evaluation. The following list briefly describes the selected benchmarks.
\begin{itemize}
    \item {MMMU~\cite{mmmu}}: The Massive Multi-discipline Multimodal Understanding benchmark consists of college-level multiple-choice and open-ended questions from 30 different disciplines. We use Chain-of-Thought (CoT) prompting for this benchmark and report accuracy.
    \item {ChartQA~\cite{chartqa}:} The 2,500 questions of this benchmark cover three different types of charts (bar, line and pie) and require strong visual, logical, and arithmetical reasoning capabilities. We evaluate on the test set and report relaxed accuracy.
    \item {DocVQA~\cite{docvqa}:} This benchmark probes capabilities on document analysis and recognition, including Optical Character Recognition (OCR). The 5,349 questions contain images from a diverse set of documents, ranging from 1940 to 2020 and covering multiple industries. We report Average Normalized Levenshtein Similarity (ANLS).
    \item {TextVQA~\cite{textvqa}:} The 5,000 samples of this dataset focus specifically on text-reading capabilities (OCR) in natural images. We report weighted accuracy on the validation set.
    \item {VATEX~\cite{vatex}:} This video captioning benchmark covers a diverse set of human activities. We evaluate on the public test set containing videos with a length of around 10 seconds. The CIDEr~\cite{cider_metric} score is used for evaluation.
    \item {EgoSchema~\cite{egoschema}:} The unique characteristic of this long-form video question answering benchmark is its high ``certificate length''~\cite{arora_computational_complexity_certificate_length}, which is, loosely speaking, the time it takes a human to verify the video description. The videos cover a broad range of natural human activities and come with human-curated multiple-choice question-answer pairs.
\end{itemize}

Table~\ref{tab:evaluations_multimodal_intelligence} summarizes our quantitative results on multiple image and video understanding benchmarks. Amazon Nova Pro and Lite achieve high scores across all benchmarks. Chart understanding on ChartQA and video understanding on VATEX stand out, where Nova models rank either first or second.
We provide the prompt templates for all benchmarks in Appendix~\ref{sec:appendix_prompts_mm}, as well as qualitative examples in Appendix~\ref{sec:appendix_evaluations_multimodal_intelligence_results}.

\subsection{Agentic workflows}

Amazon Nova Pro, Lite, and Micro models can be used as agents. An agent considers a suite of tools and APIs, reasons about the user's request and past conversational history, chooses if a tool should be used and, if so, decides which tool to use, invokes the tool, assesses the outcome from the tool, and then communicates back with the user \cite{yao2023react,schick2023toolformer,lu2023chameleon,patil2023gorillalargelanguagemodel}.

To this end, we evaluated our Nova models on agentic workflows that require textual understanding and visual reasoning.  For textual understanding (Section~\ref{sec:evaluations_agentic_workflows}), we used the Berkeley Function Calling Leaderboard benchmark to test our models' capabilities in function calling and orchestrating real-world applications.
For visual reasoning (Section~\ref{sec:evaluations_agentic_workflows_mm}), we evaluate on three benchmarks that require image understanding capabilities for correct function calling.
We highlight that both Amazon Nova Pro and Lite models set a new state of the art on these challenging benchmarks.

\subsubsection{Agentic text benchmarks and results}
\label{sec:evaluations_agentic_workflows}

Table~\ref{tab:agentic_benchmarks}  presents quantitative results on the Berkeley Function Calling Leaderboard v3 (BFCL).\footnote{BFCL is a fast-moving, live benchmark. We report results using the state of the repository and website leaderboard as of Nov 17th, 2024 (commit 8226d).} Stemming from the Gorilla project \cite{patil2023gorillalargelanguagemodel}, the revamped BFCL \cite{berkeley-function-calling-leaderboard} benchmark evaluates a model's ability to accurately call and utilize real-world functions, or tools, based on a user's natural language request. Amazon Nova models particularly excel in the Abstract Syntax Tree (AST), Execution, and Relevance metrics, as well as overall scores versus comparable models. Amazon Nova Lite and Micro also had the lowest latency of the selected models.

In Table \ref{tab:agentic_benchmarks}, AST measures the exact match function calling performance of the model when comparing function names and argument/value signatures to a human-curated ground truth. While AST allows for some soft matching based on manually-defined, permitted argument values (e.g., different date formats), Execution measures a function call's accuracy not by the call signature itself, but by comparing the return value of the call when executed against a real API. 

To measure the rate of hallucination, Irrelevance measures the model's ability to recognize that it does not have the appropriate functions available to help the user, and should therefore not call any. Relevance, as the opposite of irrelevance, measures the model's ability to recognize it indeed does have the functions necessary to help the user (but does not verify function signature accuracy). For both metrics, higher numbers are better. %

\begin{table}[ht]
\centering
\begin{tblr}{
    colspec={lX[1,c]X[1,c]X[1,c]X[1,c]X[1,c]X[1,c]X[1,c]X[1,c]},
    rows={m},
    row{1}={font=\bfseries\footnotesize},
    row{2}={font=\itshape\footnotesize},
    cell{1}{4} = {c=2}{c},
    cell{1}{8} = {c=2}{c},
}
\toprule %
 & Overall & Latency & Non-Live & & Live & Multi-Turn & Hallucination &  \\
 & accuracy ($\uparrow$) & seconds ($\downarrow$) & {AST\\($\uparrow$)} & execution ($\uparrow$) & overall ($\uparrow$) & overall ($\uparrow$) & relevance ($\uparrow$) & irrelevance ($\uparrow$) \\
\midrule
Nova Pro  & 68.4 & 1.0 & 90.1 & 89.8 & 71.5 & 45.1 & 95.1 & 65.1 \\
Nova Lite  & 66.6 & 0.6 & 87.5 & 86.4 & 66.0 & 50.3 & 97.6 & 49.1 \\
Nova Micro  & 56.2 & 0.5 & 87.2 & 89.7 & 67.4 & 15.5 & 87.8 & 57.6 \\
\midrule
Claude Sonnet 3.5 (Jun)  & 61.3 & 3.9 & 70.0 & 66.3 & 74.7 & 40.0 & 68.3 & 74.6 \\
Claude Haiku 3  & 40.4 & 1.5 & 41.7 & 47.5 & 57.7 & 20.6 & 97.6 & 29.4 \\
\midrule
Gemini 1.5 Pro (002)  & 59.8 & 3.0 & 88.0 & 91.4 & 74.3 & 16.3 & 75.6 & 75.1 \\
Gemini 1.5 Flash (002) & 55.3 & 1.1 & 79.7 & 80.6 & 73.2 & 12.5 & 78.1 & 75.7 \\
\midrule
Llama 3.2 90B$\scriptstyle ^\textit{A}$  & 54.3 & N/A & 88.9 & 89.3 & 61.1 & 14.3 & 92.7 & 58.4 \\
Llama 3.2 11B$\scriptstyle ^\textit{A}$ & 49.9 & N/A & 83.6 & 87.3 & 57.9 & 10.5 & 78.1 & 41.6 \\
\midrule
GPT-4o (Aug)  & 68.9 & 1.5 & 85.9 & 85.6 & 75.4 & 45.3 & 63.4 & 82.9 \\
GPT-4o-mini (Jul)  & 60.7 & 1.6 & 84.3 & 84.1 & 70.2 & 28.3 & 80.5 & 71.8 \\
\bottomrule %
\end{tblr}
\caption{Results on the Berkeley Function Calling Leaderboard (BFCL) v3 as of the Nov 17th, 2024 update. We include the latest versions of the models available on the leaderboard at that time. ($\scriptstyle \textit{A}$) We use leaderboard results for Llama 3.1 8B and 70B for Llama 3.2 11B and 90B, respectively, given the shared text LLM.}
\label{tab:agentic_benchmarks}
\end{table}

\subsubsection{Agentic multimodal benchmarks and results}
\label{sec:evaluations_agentic_workflows_mm}
The Amazon Nova Pro and Lite models  provide native support for multimodal inputs, including agentic workflows. In this section, we present results from our models on three different benchmarks that require agents to navigate websites to solve real-world tasks. Websites are typically represented as screenshots in these datasets to correctly convey all style elements and visual data as rendered in a standard web browser.
\begin{itemize}
    \item {VisualWebBench~\cite{visualwebbench}}: This benchmark includes seven core tasks related to web browsing, including captioning, question answering, OCR, action prediction, and grounding. All models are evaluated on 1,536 samples that span more than 100 websites from 12 domains. The final metric is the average over different metrics for the individual core tasks.
    \item {MM-Mind2Web~\cite{mind2web-mm}}: This extension of the original Mind2Web~\cite{mind2web} benchmark links samples with the original website screenshots, making it multimodal. 
    An agent needs to select an element and pick one of three elementary actions (click, type, or select) alongside a value for some actions. We report micro average over the per-sample step accuracy, where an agent is successful only if element and action selection, as well as the predicted value, are correct.
    \item {GroundUI-1K~\cite{groundingui}}: This benchmark is composed of multiple existing datasets, including Mind2Web~\cite{mind2web}, and repurposes them as a grounding task. On 1,000 samples for evaluation, a multimodal agent is given an instruction and a screenshot of a website from a wide variety of domains and asked to predict the 2D location of the desired UI element. The agent is correct if its predicted 2D location is within the ground truth bounding box.
\end{itemize}

Table~\ref{tab:evaluations_agentic_workflows_mm_results} shows the results of our models on multimodal agent workflows along with other publicly-reported results.
Both Amazon Nova models, Lite  and Pro, demonstrate strong visual reasoning and agentic capabilities and achieve high scores on all three benchmarks. 

\begin{table}[th]
\centering
\begin{tblr}{
    width=0.8\textwidth,
    colspec={lX[1,c]X[1,c]X[1,c]},
    rows={m},
    row{1}={font=\bfseries},
    row{2}={font=\itshape}
}
\toprule
    & VisualWebBench & MM-Mind2Web & GroundUI-1K \\
    & composite$\scriptstyle ^\textit{D}$  & step accuracy & accuracy \\
\midrule
Nova Pro  & 79.7 & 63.7 & 81.4 \\
Nova Lite & 77.7 & 60.7 & 80.2 \\
\midrule
Claude 3.5 Sonnet (Oct)   & 76.7\TblrNote{$\scriptstyle \textit{M}$} & 61.6\TblrNote{$\scriptstyle \textit{M}$} & 16.3 \\
\midrule
GPT-4o (Nov) & 77.5\TblrNote{$\scriptstyle \textit{M}$} & 55.0\TblrNote{$\scriptstyle \textit{M}$} & 13.4\TblrNote{$\scriptstyle \textit{C}$} \\
GPT-4o Mini (Jul) & 71.3\TblrNote{$\scriptstyle \textit{M}$} & 58.6\TblrNote{$\scriptstyle \textit{M}$} & 7.2\TblrNote{$\scriptstyle \textit{M}$} \\
GPT-4 (Apr)            & 64.6 & 36.8\TblrNote{$\scriptstyle \textit{A}$} & - \\
\midrule
Gemini 1.5 Pro (002) & 76.4\TblrNote{$\scriptstyle \textit{M}$} & 58.4\TblrNote{$\scriptstyle \textit{M}$} & 35.2\TblrNote{$\scriptstyle \textit{B}$} \\
Gemini 1.5 Flash (002) & 76.1\TblrNote{$\scriptstyle \textit{M}$} & 46.2\TblrNote{$\scriptstyle \textit{M}$} & 59.9\TblrNote{$\scriptstyle \textit{M}$} \\
Gemini 1.0 Pro (001) & 48.0 & 17.9\TblrNote{$\scriptstyle \textit{A}$} & - \\
\midrule
Llama 3.2 90B  & 73.2\TblrNote{$\scriptstyle \textit{M}$} & 21.6\TblrNote{$\scriptstyle \textit{M}$} & 8.3\TblrNote{$\scriptstyle \textit{M}$} \\
Llama 3.2 11B  & 65.1\TblrNote{$\scriptstyle \textit{M}$} & 22.1\TblrNote{$\scriptstyle \textit{M}$} & 3.7\TblrNote{$\scriptstyle \textit{M}$} \\
\bottomrule
\end{tblr}
\caption{
Quantitative results on three multi-modal agentic benchmarks: VisualWebBench~\cite{visualwebbench}, MM-Mind2Web~\cite{mind2web-mm} and GroundUI-1K~\cite{groundingui}. Reference numbers are taken from the corresponding benchmark papers~\cite{visualwebbench,mind2web-mm,groundingui} and leaderboard~\cite{website_groundingui_leaderboard}.
Remarks:
($\scriptstyle \textit{A}$) uses in-context learning (ICL) (please note that Amazon Nova models do not need to rely on in-context examples);
($\scriptstyle \textit{B}$) Gemini 1.5 Pro (001);
($\scriptstyle \textit{C}$) GPT-4o (May);
($\scriptstyle \textit{D}$) Macro average over individual metrics;
($\scriptstyle \textit{M}$) Measured by us.\textsuperscript{\ref{api_measurements}}
}
\label{tab:evaluations_agentic_workflows_mm_results}
\end{table}

\subsection{Long context}
\label{sect:eval_longcontext}
We evaluate Amazon Nova Pro, Lite, and Micro on tasks that require the models to understand and reason over long context. These skills are crucial for tasks such as long multi-turn conversations, reasoning over long lists of retrieved documents, or understanding long videos. Amazon Nova Micro, Lite, and Pro models support context lengths of 128k, 300k, and 300k tokens, respectively.
We used the following benchmarks to evaluate our models' long context performance:

\begin{itemize}
    \item {Text Needle-in-a-Haystack (NIAH):} Following~\cite{kamradt2023}, we assessed each model's ability to locate specific information (the ``needle'') within extensive contexts (the ``haystack''). This ``needle-in-a-haystack'' test evaluates the model's performance on context lengths starting at 32k, allowing us to measure its ability to accurately retrieve information across varying lengths of input context. 
    \item {SQuALITY~\cite{wang-etal-2022-squality} (ZeroScrolls Benchmark~\cite{shaham-etal-2023-zeroscrolls}):} Focused on query-based summarization of literary stories, this task evaluates the model's capacity to generate relevant summaries from large contexts.
    \item {LVBench \cite{wang2024lvbench}:} This multimodal benchmark includes questions about YouTube videos\footnote{\url{https://huggingface.co/datasets/AIWinter/LVBench}} from various domains such as TV series, sports, broadcasts, and surveillance footage. The LVBench dataset consists of 99 videos and 1,549 questions, covering six different types of tasks such as reasoning, event understanding and summarization.

\end{itemize}
\begin{figure}[th]
    \centering
    \includegraphics[width=1.0\linewidth]{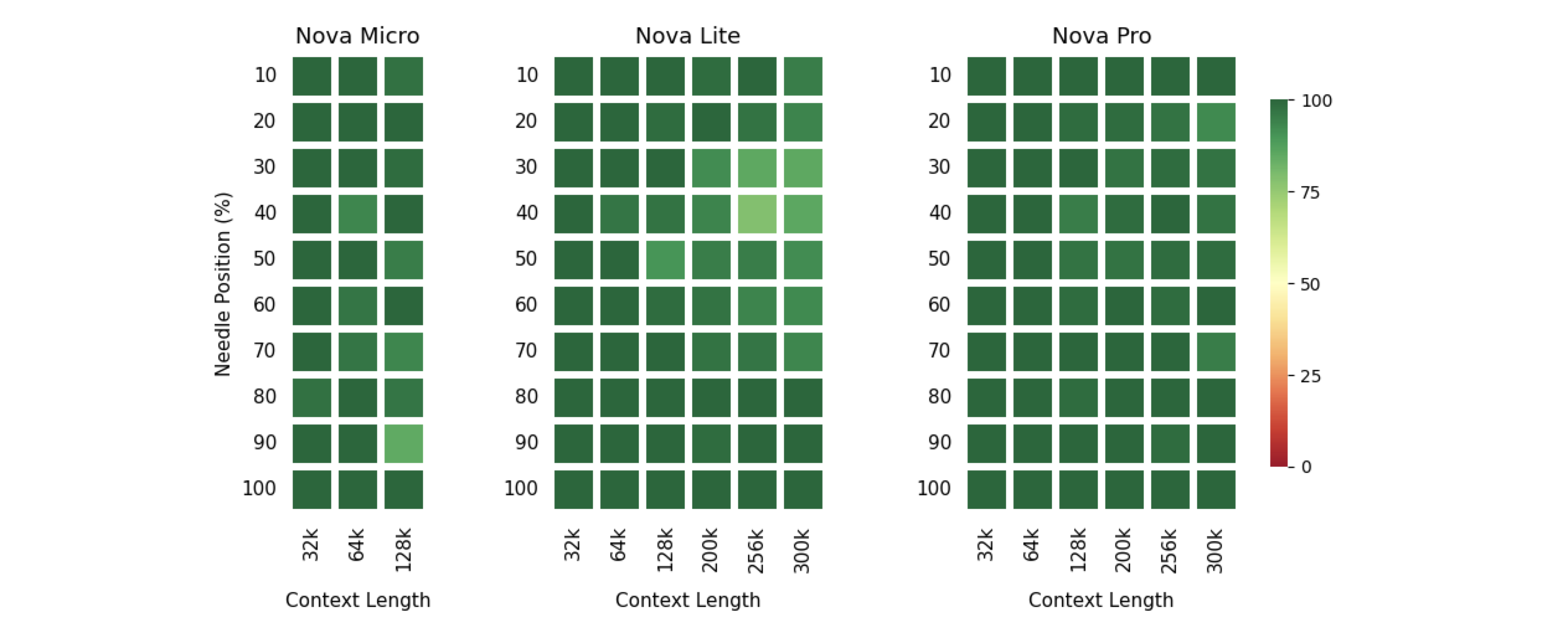}
    \caption{Text Needle-in-a-Haystack recall performance for Nova Micro (up-to 128k), Nova Lite (up-to 300k) and  Nova Pro (up-to 300k) models.}
    \label{fig:niah-recall-grid}
\end{figure}

\begin{table}[th]
\centering
\begin{tblr}{
    width=0.6\textwidth,
    colspec={lX[1,c]X[1,c]},
    rows={m},
    row{1}={font=\bfseries},
    row{2}={font=\itshape},
}
\toprule
& SQuALITY & LVBench \\
& ROUGE-L & accuracy \\
\midrule
Nova Pro         & 19.8\ci{8.7} & 41.6\ci{2.5}      \\
Nova Lite        & 19.2\ci{8.6} & 40.4\ci{2.4}      \\
Nova Micro       & 18.8\ci{8.6} & -      \\
\midrule
Claude 3.5 Sonnet (Jun) & 13.4\ci{7.5} & - \\
\midrule
Gemini 1.5 Pro (001) & - & 33.1\ci{2.3} \\ %
Gemini 1.5 Pro (002)  & 19.1\ci{8.6}\TblrNote{$\scriptstyle \textit{M}$} & - \\
Gemini 1.5 Flash (002) & 18.1\ci{8.4}\TblrNote{$\scriptstyle \textit{M}$} & - \\
\midrule
GPT-4o           & 18.8\ci{8.6} & 30.8\ci{2.3} \\ %
\midrule
Llama 3 - 70B    & 16.4\ci{8.1} & - \\
Llama 3 - 8B     & 15.3\ci{7.9} & - \\
\bottomrule
\end{tblr}
\caption{
Text and Multimodal long context performance on SQuALITY (ROUGE-L) and LVBench (Accuracy). For SQuALITY, measurements for Claude 3.5 Sonnet, GPT-4o, Llama 3 70B and Llama 3 8B are taken from the Llama 3 report \cite{llama31}. Gemini results were measured by us\textsuperscript{\ref{api_measurements}} ($\scriptstyle \textit{M}$). For LVBench, Gemini and GPT-4o numbers were taken from the corresponding benchmark leaderboard~\cite{wang2024lvbench}.}
\label{tab:long_context_text}
\end{table}

Results for text and multimodal long context benchmarks are presented in Table~\ref{tab:long_context_text}. In the long video question answering task, both Amazon Nova Pro and Lite demonstrate robust performance on the LVBench dataset, surpassing other models. Amazon Nova models consistently demonstrate exceptional performance in retrieving information from any depth across both text and multimodal understanding use cases, delivering high accuracy and reliability.

\subsection{Functional expertise}

In addition to core capabilities, foundation models must perform well in particular specialties and domains. Across our many areas of performance analyses, we have selected four domains for which to present benchmarking results: Software engineering, financial analysis, and retrieval-augmented generation.
Prompt templates for all benchmarks can be found in Appendix \ref{sec:prompt-vertical}.

\begin{table}[ht]
\centering
\begin{tblr}{
    width=0.8\textwidth,
    colspec={lX[-1,c]X[1,c]X[1,c]X[1,c]},
    column{2}={font=\footnotesize},
    rows={m},
    row{1}={font=\bfseries},
    row{3}={font=\itshape},
}
\toprule
 &           & Software & Finance & RAG \\
 & & {HumanEval\\Python} & {FinQA} & CRAG \\
 & {tok/\\sec} & {0-shot\\pass@1} & {0-shot\\accuracy} & accuracy \\
\midrule
Nova Pro  	             & 100 & 89.0\ci{4.8} 	 & 77.2\ci{0.9} & 50.3\ci{1.9} \\
Nova Lite  		         & 157 & 85.4\ci{5.4} 	 & 73.6\ci{0.9} & 43.8\ci{1.9} \\
Nova Micro 			     & 210 & 81.1\ci{6.0}	 & 65.2\ci{1.0} & 43.1\ci{1.9} \\
\midrule
Claude 3.5 Sonnet (Oct)  	& 57    & 93.7\ci{3.7}		& 77.3\ci{0.9}\TblrNote{$\scriptstyle \textit{M}$} & 52.6\ci{1.8}\TblrNote{$\scriptstyle \textit{M}$} 		\\
Claude 3.5 Haiku  	& 64 & 88.1\ci{5.0}	 & 73.9\ci{0.9}\TblrNote{$\scriptstyle \textit{M}$} & 31.9\ci{1.8}\TblrNote{$\scriptstyle \textit{M}$} \\
\midrule
Gemini 1.5 Pro (002) 				& 58     & 87.8\ci{5.0}\TblrNote{$\scriptstyle \textit{M}$}	 & 74.4\ci{0.9}\TblrNote{$\scriptstyle \textit{M}$} & 48.9\ci{1.9}\TblrNote{$\scriptstyle \textit{M}$}\\
Gemini 1.5 Flash (002) 			& 190   & 81.1\ci{6.0}\TblrNote{$\scriptstyle \textit{M}$} & 73.5\ci{1.0}\TblrNote{$\scriptstyle \textit{M}$} & 42.4\ci{1.9}\TblrNote{$\scriptstyle \textit{M}$}		\\
Gemini 1.5 Flash 8B (001) & 283 & 81.1\ci{6.0}\TblrNote{$\scriptstyle \textit{M}$}  & 63.7\ci{1.0}\TblrNote{$\scriptstyle \textit{M}$} & 37.7\ci{1.8}\TblrNote{$\scriptstyle \textit{M}$} \\
\midrule
GPT-4o 					& 163    & 90.2\ci{4.6}	 & 71.1\ci{1.0}\TblrNote{$\scriptstyle \textit{M}$} & 52.0\ci{1.9}\TblrNote{$\scriptstyle \textit{M}$}\\
GPT-4o Mini 				& 113   & 87.2\ci{5.1}	 & 70.6\ci{1.0}\TblrNote{$\scriptstyle \textit{M}$} & 49.9\ci{1.9}\TblrNote{$\scriptstyle \textit{M}$} \\
\midrule
Llama 3.2 90B & 40 & 80.5\ci{6.1} & 72.8\ci{1.0}\TblrNote{$\scriptstyle \textit{M}$} & 45.2\ci{1.9}\TblrNote{$\scriptstyle \textit{M}$} \\
Llama 3.2 11B & 124 & 72.6\ci{6.8} & 60.8\ci{1.1}\TblrNote{$\scriptstyle \textit{M}$} & 42.2\ci{1.9}\TblrNote{$\scriptstyle \textit{M}$}\\
Llama 3.1 8B 					& 157 & 72.6\ci{6.8} & 61.2\ci{1.0}\TblrNote{$\scriptstyle \textit{M}$} & 42.2\ci{1.8}\TblrNote{$\scriptstyle \textit{M}$} \\
\bottomrule
\end{tblr}
\caption{
Performance on select functional benchmarks, including software engineering benchmarks in Python with HumanEval~\cite{chen2021evaluating}, financial reasoning with FinQA~\cite{Chen2021FinQAAD}, and retrieval augmented generation with CRAG~\cite{yang2024crag}. CRAG uses our scoring method described in Section~\ref{ssec:rag}. Where available, reference numbers are taken from the corresponding benchmark papers and technical reports~\cite{anthropic2024claude, claude3modelfamily, gemini_1_5, islam2024gpt, llama31, o1mini}. Additional results were measured ($\scriptstyle \textit{M}$) by us\textsuperscript{\ref{api_measurements}}. Model speed in tokens per second (Tok/Sec) is reproduced from section \ref{sec:eval_runtime}.
}
\label{tab:functional}

\end{table}

\subsubsection{Software engineering}

We assessed Amazon Nova's code generation capabilities on the Python coding task HumanEval~\cite{chen2021evaluating}.  
The benchmark contains 164 original programming problems with unit tests. These problems assess language comprehension, algorithms, and simple mathematics. Some problems are comparable to simple software interview questions. 
Table~\ref{tab:functional} provides the performance of our Nova models and select public models.

\subsubsection{Financial analysis}

We use FinQA \cite{Chen2021FinQAAD} to evaluate Amazon Nova's ability to understand financial data. FinQA is an expert-annotated dataset comprising 8,281 financial question-answer pairs derived from the earnings reports of S\&P 500 companies. It evaluates a model's ability to extract information from both tables and unstructured text, while accurately performing calculations using relevant financial knowledge. We report the average post-rounding accuracy under the 0-shot CoT setting. Table~\ref{tab:functional} provides the performance of Amazon Nova models and select public models on FinQA.

\subsubsection{Retrieval augmented generation}
\label{ssec:rag}

We evaluate RAG capabilities on the CRAG~\cite{yang2024crag} benchmark using the Task 1 setup, which considers five pre-selected HTML pages as external knowledge to each input question. We extract top-20 text snippets from these pages following the standard retrieval approach used in CRAG's official repository, whereby pages are first cleaned using BeautifulSoup to remove HTML tags, after which the text is then split into sentences or chunks no longer than 1000 characters. These are then encoded using the \textit{sentence-transformers/all-MiniLM-L6-v2} model, which is also used to encode the question. The top 20 chunks with highest similarity are passed as context in the input for model inference.
We report the percentage of correct responses as judged by an LLM (\textit{gpt-4-turbo-2024-04-09}), which compares each model's answer with the expected answer using the prompt shown in Appendix \ref{sec:prompt-rag-eval}.
Table~\ref{tab:functional} provides the performance of Amazon Nova models and selected public models on a combined validation and test set of 2,706 examples.

\subsection{Runtime performance}
\label{sec:eval_runtime}

We evaluate the runtime performance of Amazon Nova models using three metrics: Time to First Token (TTFT), Output Tokens per Second (OTPS) and Total Response Time. TTFT is measured as the time, in seconds, it takes to receive the first token from the model after an API request is sent. OTPS is measured as the number of tokens generated per second (tok/sec). It is the rate at which a model produces subsequent output tokens after the first token, reflecting overall throughput and efficiency during inference. Total Response Time measures the total duration in seconds from the submission of the input prompt to the end of generation sequence for a given input-output prompt length. It represents the overall user experience for a model.

\begin{figure}[p]
    \centering
    \includegraphics[width=0.9\textwidth]{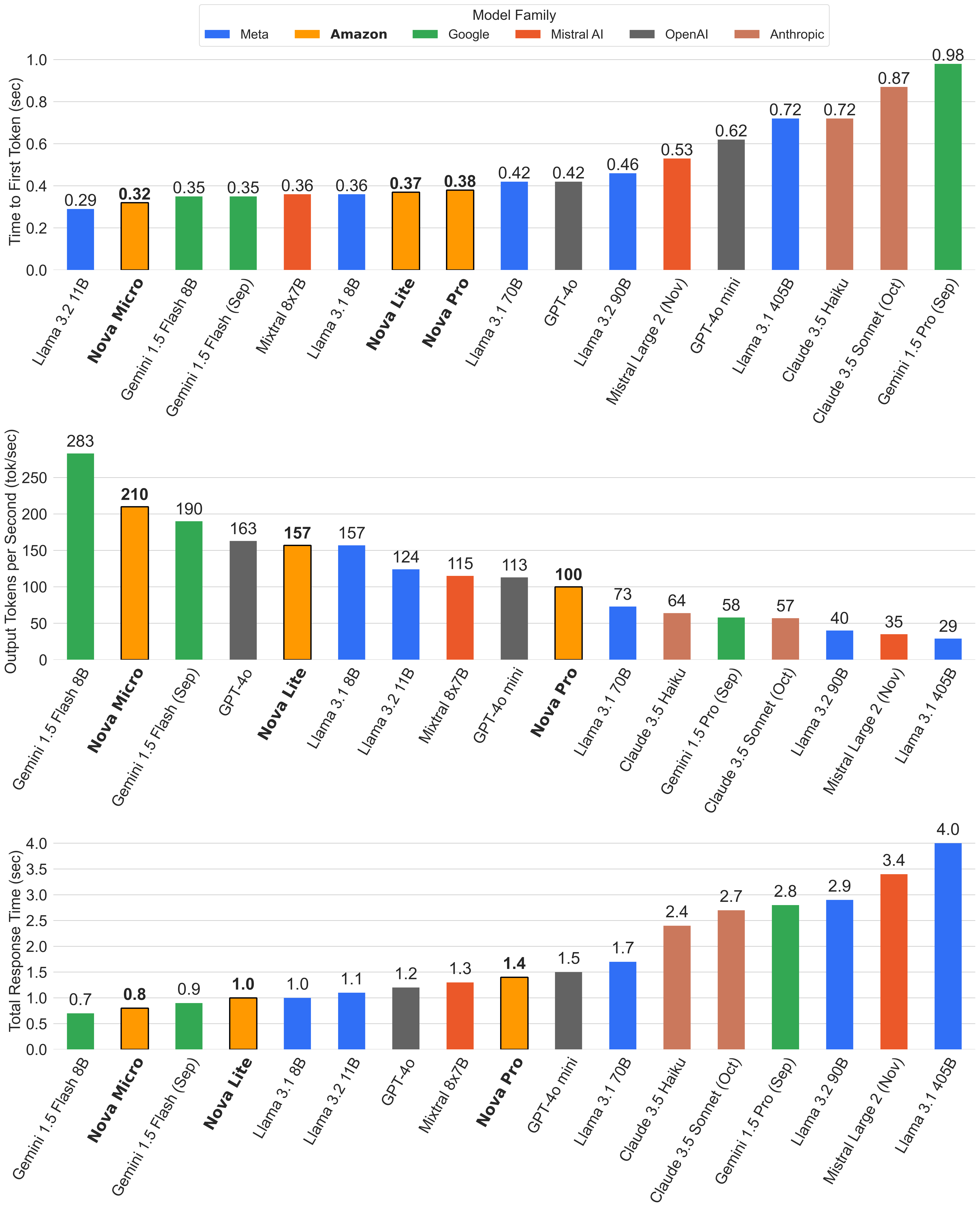}
    \caption{Time to First Token ($\downarrow$), Output Tokens per Second ($\uparrow$), and Total Response Time ($\downarrow$) using 1,000 tokens of input and 100 tokens of output for Amazon Nova models and select publicly-available models (Artificial Analysis, Nov 29th, 2024).}
    \label{fig:runtime_performance}
\end{figure}

In Figure \ref{fig:runtime_performance}, we show TTFT, OTPS, and Total Response Time using 1000 tokens of input and 100 tokens of output for Amazon Nova models and select public models as reported by Artificial Analysis\footnote{\url{https://artificialanalysis.ai/methodology}}, an independent entity that benchmarks AI models and hosting providers. 
Amazon Nova Micro, Lite and Pro models are among the fastest models in their respective intelligence tiers. Together, all three Amazon Nova models demonstrate state-of-the-art runtime performance, ensuring a smooth and responsive user experience in many real world use cases.

\clearpage
\section{Amazon Nova Canvas Evaluation}
Amazon Nova Canvas is a diffusion model that takes a text prompt and an optional RGB image as input and generates an image as an output conditioned on the input text and optional image. Illustrative examples of the images generated by Amazon Nova Canvas can be found in our Amazon Science blog post~\footnote{~\url{https://www.amazon.science/blog/amazon-nova-canvas-examples}}. In this section, we provide details on the evaluation strategy and performance of the model both in terms of automated metrics and human evaluation.

\subsection{Automated metrics}
We use ImageReward~\cite{xu2024imagereward} and Text-to-Image Faithfulness (TIFA)~\cite{hu2023tifa} as automated metrics. 

\begin{itemize}
    \item ImageReward score is generated from a standardized reward model that aligns human preference with the predicted score. To compute the ImageReward score, we randomly sample 10k prompts from MSCOCO-2014~\cite{lin2014microsoft} validation set and use this set for calculating the score.
    \item Text-to-Image Faithfulness (TIFA) score is a reference-free metric that measures the faithfulness of a generated image to the input text via visual question answering (VQA). The evaluation set for TIFA score is a pre-selected 4k prompts in the TIFA-v1.0 benchmark, sampled from MSCOCO captions~\cite{lin2014microsoft}, DrawBench~\cite{saharia2022photorealistic}, PartiPrompts~\cite{yu2022scaling}, and PaintSkill~\cite{cho2023dall} datasets. 
\end{itemize}

We compare Amazon Nova Canvas with other publicly-available models including DALL.E 3~\cite{betker2023improving}, Stable Diffusion 3 Medium ~\cite{esser2024scaling}, Stable Diffusion 3.5 Large~\cite{sd3.5} and Flux (Schnell and Pro)~\cite{flux2024}. The results are shown in Table~\ref{tab: tifa_imagereward_oig}. 

\begin{table}[ht!]
\centering
\begin{tblr}{
  width = 0.6\textwidth,
  colspec = {lcc},
  row{1} = {font=\bfseries},
}
\toprule
   &TIFA &ImageReward \\
\midrule
Amazon Nova Canvas &  \textbf{0.897} &  \textbf{1.250} \\
DALL.E 3 &0.863 &1.052 \\
Stable Diffusion 3.5 Large & 0.891 & 1.082 \\
Stable Diffusion 3 Medium  &0.881 &0.952 \\
Flux Pro 1.0 & 0.875 &1.075 \\
Flux Schnell & 0.882 &0.999 \\
\bottomrule
\end{tblr}
\caption{Comparison of TIFA and ImageReward metrics of Amazon Nova Canvas with other models.}
\label{tab: tifa_imagereward_oig}
\end{table}

\subsection{Human evaluation}
We conduct A/B testing to compare Amazon Nova Canvas with other third-party text-to-image models. The A/B testing prompt set is composed of approximately 1,000 prompts designed to capture customer usage of text-to-image models. This set include prompts from  datasets such as MSCOCO~\cite{lin2014microsoft}, Drawbench~\cite{saharia2022photorealistic}, OpenParti~\cite{yu2022scaling}, DALL.E 3 Eval~\cite{betker2023improving}, and DOCCI~\cite{onoe2404docci} and covers a broad set of categories such as humans, landscapes, natural scenarios, indoor environments, creative themes, artistic themes, and so forth. A few prompts were randomly selected and repeated in order to get additional data points on the quality of the model.

With each prompt we generate an image from Amazon Nova Canvas as well as each other text-to-image model. We used random seeds to generate the images from Amazon Nova Canvas and all images were generated at 1k x 1k resolution.  If the prompts trigger filters such that an image is not generated, for either the Amazon Nova Canvas model or the public text-to-image model, we ignore that prompt and do not show it to the human raters. All human evaluation is done in a single-blind manner where the annotator is provided two sets of images, one from Amazon Nova Canvas and the other from the third-party model. The order of the images are randomized for each prompt and annotator. In our blind testing, we ask human annotators to select images that they prefer based on (1) text-image alignment, which measures the instruction-following capability of the model, and (2) image quality, which quantifies the overall preference of the annotators. To ensure rigorous, consistent, and unbiased evaluation, we used a third-party vendor for human evaluation. We created guidelines that were used to train the annotators so that the decision-making criteria were clear to them in each dimension. 

The pair-wise results comparing Amazon Nova Canvas with OpenAI DALL.E 3 and Google Imagen 3 are shown in Table~\ref{tab:oig_ab_results}, including win, tie, loss rate. The win rate reflects the percentage of samples where Amazon Nova Canvas was preferred over the other model while the tie rate indicates the scenario where the human annotator did not perceive a difference between the two models. As can be seen in the results, Amazon Nova Canvas has a higher win rate compared to the other text-to-image models.

\begin{table}[ht!]
\centering
\begin{tblr}{
  width = 0.8\textwidth,
  colspec = {lcccccc},
  row{1} = {font=\bfseries},
  row{2} = {font=\itshape},
}
\toprule
Nova Canvas versus: & \SetCell[c=3]{c} DALL.E 3 & & & \SetCell[c=3]{c} Imagen 3 & & \\
 & win rate & tie rate & loss rate & win rate & tie rate & loss rate \\
\midrule
Overall preference (image quality) & 54.5 & 6.4 & 39.1 & 48.2 & 5.3 & 46.5 \\
Instruction following (text-image alignment) & 39.4 & 22.5 & 38.1 & 38.4 & 28.1 & 33.5 \\
\bottomrule
\end{tblr}
\caption{The win, tie, and loss rates (\%) from human evaluation of Amazon Nova Canvas versus (a) DALL.E 3 and (b) Imagen 3.}
\label{tab:oig_ab_results}
\end{table}

\section{Amazon Nova Reel Evaluation}
Amazon Nova Reel is a diffusion model that takes a text prompt and an optional RGB image as input and generates a video as an output conditioned on the input text and optional image. Illustrative examples of the videos generated by the Amazon Nova Reel can be found in our Amazon Science blog post.\footnote{\url{https://www.amazon.science/blog/amazon-nova-reel-examples}} In this section, we provide details on the evaluation strategy and performance of the model.

\subsection{Human evaluation metrics}

To evaluate Amazon Nova Reel, we rely on human feedback to assess the generated videos across two primary axes: video quality and video consistency. All evaluations are conducted through single-blind pairwise comparisons. Human annotators are provided a set of two videos shown side-by-side and are asked to choose the better video or mark them as equal if they find the videos to be equally performant across the metric on which they are evaluating. All videos were generated in 720p resolution and different random seeds were used during generation.

The \textit{video quality} axis encapsulates the technical and perceptual aspects of the generated video via four primary components: 
\begin{itemize}
    \item Image quality: The visual appeal of individual frames, including resolution, sharpness, object clarity, and overall composition, where each frame is visually pleasing and artifact-free.
    \item Motion quality: The fluidity of movement across frames, including motion consistency and smooth transitions without flickering, distortion, or abrupt shifts, contributing to natural and realistic motion portrayal.
    \item Image-text alignment: How closely individual frames match the prompt, considering the presence of described entities, their attributes, spatial relationships, colors, and other static visual details.
    \item Motion-text alignment: The accuracy of dynamic elements, including the correctness of actions performed by entities, camera movements, and temporal changes in attributes, as well as adherence to the provided description.
\end{itemize}
The video quality axis additionally includes factors influencing overall appeal, such as motion degree, entity size, creative composition, and general video likability. 

The \textit{video consistency} axis encapsulates the temporal coherence of both subjects and backgrounds throughout the video. It includes assessments of the maintenance of entity size, shape, and appearance, as well as background stability without unexpected morphing or changes. A high score in this dimension means believable spatial relationships between foreground and background elements throughout the video duration. 

In combination, the video quality and video consistency metrics provide a holistic and robust evaluation framework for video generation models by considering both technical accuracy and perceptual appeal.

\subsection{Dataset}
We curated a diverse set of prompts designed to capture various aspects of video generation. The prompts are distributed across 6 broad categories: human and activities, animals, natural scenery and landscapes, indoor scenes, objects interactions, and creative scenes and activities. This broad categorization ensures that the evaluation covers a wide range of real-world scenarios. We structured the prompt set to cover various motion-related aspects, which is critical for assessing motion-text alignment in the generated videos. For example, we included prompts with a variety of camera motions to evaluate how well the models follow instructions related to camera movement. Additionally, we incorporated dynamic attributes~\cite{sun2024t2v}, in which the subject or background undergoes state or shape changes over time, which allows us to evaluate the model’s ability to generate evolving entities. Finally, we added prompts that require motion binding~\cite{sun2024t2v}, where specific compositions of movements and actions are requested, enabling us to assess how well models can generate complex, coordinated motions. The curated prompt set consists of approximately 700 prompts, all from various open source benchmarks.

\subsection{Implementation details \& results}
To ensure a rigorous, consistent and unbiased evaluation process, we outsourced the annotation collection process to a third-party vendor. We created detailed guidelines, in which annotators were given comprehensive instructions and examples for each evaluation dimension, ensuring clarity on the criteria for marking preferences between videos. These guidelines included examples of different scenarios to aid in decision-making across our evaluation axes. Alongside this, we ensured that annotators were trained using expert-provided examples, with each round of annotations subject to spot checks. Specifically, 5-10\% of the data from each batch was randomly selected and reviewed by expert annotators. Based on this feedback, the vendor continuously refined the annotators' understanding and accuracy, ensuring a high standard of evaluation across the board. To further enhance the reliability of the results, we employed a consensus voting system. For each video comparison, annotations were collected from three different evaluators, and a majority voting approach was used to determine the final outcome. This method helps reduce individual biases and ensures that the final assessments are based on collective judgment, thereby increasing the robustness of the evaluation.

For reporting performance, we conducted pairwise comparisons between Amazon Nova Reel and other state-of-the-art models including Gen3 Alpha~\cite{Gen3-alpha} by Runway ML and Luma 1.6~\cite{Luma1.6} by Luma Labs. We report results in terms of win,  tie, and loss rates. The win rate reflects the percentage of samples where Amazon Nova Reel was preferred over the other model, while the tie rate indicates cases where no perceptible difference between the two models was found by the evaluators. Using the curated prompt set described earlier, we evaluate the models across all the dimensions outlined above, and report the results in Table~\ref{tab:ovg_results}.

\begin{table}[ht]
\centering
\begin{tblr}{
  width = 0.8\textwidth,
  colspec = {lcccccc},
  row{1} = {font=\bfseries},
  row{2} = {font=\itshape},
}
\toprule
Nova Reel versus: & \SetCell[c=3]{c} Runway Gen3 Alpha & & & \SetCell[c=3]{c} Luma 1.6 & & \\
& win rate & tie rate & loss rate & win rate & tie rate & loss rate \\
\midrule
Video Quality & 56.4 & 9.9 & 33.7 & 51.1 & 3.4 & 45.5 \\
Video Consistency & 67.0 & 9.1 & 23.9 & 74.7 & 5.1 & 20.2 \\
\bottomrule
\end{tblr}
\caption{The win, tie, and loss rates (\%) from human evaluation of Amazon Nova Reel versus (a) Gen3-Alpha and (b) Luma1.6.}
\label{tab:ovg_results}
\end{table}

In video consistency, Amazon Nova Reel achieved win rates of 67.0\% against Gen3 Alpha and 74.7\% against Luma 1.6, demonstrating superior subject and background coherence. For video quality, Amazon Nova Reel secured win rates of 56.4\% against Gen3 Alpha and 51.1\% against Luma 1.6.

\section{Responsible AI}

Our approach to Responsible AI (RAI) is structured around eight foundational dimensions \cite{website_amazon_rai_dimensions} shown in Table~\ref{tab:ai_concepts}. 
These dimensions guide our approach to RAI for the Amazon Nova family of models, which we articulate in the following three sections: (1) defining our RAI design objectives, (2) our actions to ensure adherence to these objectives, and (3) system evaluation and red teaming. The last two components form a continuous loop of model development and human/automated verification to ensure that our Amazon Nova models are aligned with our RAI objectives and deliver an exceptional and delightful customer experience.

\begin{table}[h!]
\centering
\begin{tblr}{
    colspec={X[1,l]X[3,l]}
}
\toprule
\textbf{Term} & \textbf{Definition} \\
\midrule
Fairness & Considering impacts on different groups of stakeholders \\
Explainability & Understanding and evaluating system outputs \\
Privacy and security & Appropriately obtaining, using, and protecting data and models \\
Safety & Preventing harmful system output and misuse \\
Controllability & Having mechanisms to monitor and steer AI system behavior \\
Veracity and robustness & Achieving correct system outputs, even with unexpected or adversarial inputs \\
Governance & Incorporating best practices into the AI supply chain, including providers and deployers \\
Transparency & Enabling stakeholders to make informed choices about their engagement with an AI system \\
\bottomrule
\end{tblr}
\caption{Our eight core Responsible AI dimensions}
\label{tab:ai_concepts}

\end{table}

\subsection{Defining our RAI objectives} 
We operationalize our RAI dimensions into a series of detailed design objectives that guide our decision-making throughout the entire model development lifecycle, from initial data collection and pre-training to the implementation of post-deployment runtime mitigations. 

In addition to being grounded on the RAI dimensions, our objectives are informed by relevant laws and regulations, voluntary frameworks, and our commitments to our customers, and they undergo an internal alignment process that includes reviews from a number of stakeholders. We will continue to iterate on these objections as we engage with external experts and participate in industry and government forums, including the Frontier Model Forum \cite{blog_frontier_model_forum}, Partnership on AI \cite{blog_amazon_partnership_on_ai}, and various forums organized by government agencies such as the National Institute of Standards and Technology (NIST) of the U.S. Department of Commerce \cite{blog_amazon_nist}.

\paragraph{Our commitment to Responsible Scaling:}
As the capabilities of AI models increase (through increased training data, model size or architecture innovations), so do the potential risks that they present. We joined other technology companies in signing on to the White House's voluntary commitments on the safe, secure, and transparent development and use of foundation models \cite{blog_amazon_rai_commitment}. Since then we have actively participated in other efforts, including the AI Safety Summits in the UK and Seoul, and we have committed to new standards like the G7 AI Hiroshima Process Code of Conduct \cite{hiroshima_code_of_conduct} in accordance with our commitment to the US White House on ensuring Safe, Secure, and Trustworthy Development and Use of Artificial Intelligence. We also started a partnership with the Model Evaluation and Threat Research (METR) center\footnote{\url{https://metr.org/}} to enrich our \textit{Controllability} design objectives.

\subsection{Ensuring adherence to RAI objectives}
We employed a number of methods to measure and ensure compliance for each of our core RAI dimensions depending on their scope (i.e., whether they apply to model output, data management or other processes). For the dimensions that govern model behavior (\textit{Safety}, \textit{Fairness}, \textit{Veracity and Robustness}, \textit{Controllability}, and \textit{Privacy and Security}), we curated the pre-training data and we used both Supervised Fine Tuning (SFT) and Reinforcement Learning from Human Feedback (RLHF) methods to align our models. Based on the objectives for each RAI dimension, we created single- and multi-turn RAI demonstrations in multiple languages and conducted helpfulness/harmfulness studies to decide on SFT data mixes. We collected human preference data to be used as inputs to RLHF training where we also provided an RAI-specific reward model. We also identify risk areas during our offline evaluation or red teaming exercises (Section \ref{sec:rai-qual-eval}) and collect semantically similar examples to be included in future SFT and RLHF rounds. 

In addition to the RAI model alignment, we built runtime input and output moderation models which serve as a first and last line of defense and allow us to respond more quickly to newly identified threats or gaps in model alignment. The main role of the input moderation model is to detect prompts that contain malicious, insecure or illegal material, or attempt to bypass the core model alignment (prompt injection, jailbreaking).  Similarly, the output moderation ensures that the content adheres to our RAI objectives.

We have a rigorous \textit{Governance} methodology, developing our models in a working-backwards product process that incorporates RAI at the design phase, design consultations and implementation assessments by dedicated RAI science and data experts, and includes routine testing, reviews with customers, best practice development, dissemination, and training.

We work to ensure that our \textit{Privacy and Security} objectives are adhered to for both the model and training data. In addition to the model output alignment described above, we take measures that include data access controls \cite{website_amazon_data_protection} protecting our model training data, resulting weights, and model versions, and watermarking model outputs (see below). We address the latter through several layers of defense, including de-identifying or removing certain types of personal data from our training data, when feasible, as well as evaluation through red teaming exercises that cover data privacy assessments.

For \textit{Explainability} of our models' outputs we conduct and leverage the current active research in the area of Explainable AI to deeply understand our models' current behavior, their potential future behavior, and to build capabilities to continuously correct their behavior as and when necessary. We use various explainable AI methods throughout our model development to guide our decisions regarding RAI alignment and other mitigations. Services like Clarify \cite{website_amazon_clarify} also enable our downstream developers to easily explain model predictions.   

To work to ensure our models' \textit{Robustness} against adversarial inputs such as those that attempt to bypass alignment guardrails, we focused on risks applicable to both developers building applications using our models, and users interacting with our models via those applications. We organized those risks in broad categories such as sensitive data exfiltration, execution of unauthorized action, degradation of run-time model service availability, and malicious content generation. We used this risk organization to build model resiliency against interactions that lead to the prioritized risks.

Finally, to maximize \textit{Transparency}, we incorporate an invisible watermark during the image or video generation process and add C2PA\footnote{\url{https://c2pa.org/}} metadata in all Canvas generated content. We enhanced the robustness to alterations like rotation, resizing, color inversion, and flipping. For videos, we embed our watermark in each frame and ensure that our watermarking and detection methods withstand H264 compression. To enable anyone to easily detect the watermarks in Amazon Nova generated content, an API will be available soon after launch. Our watermark detection system introduces several enhancements such as making confidence score-based predictions instead of a single binary prediction that reflects the extent to which the generated content has been edited even when using external tools. The new detection system covers both images and videos.

\subsection{RAI Evaluation} \label{sec:rai-quant-eval}
Throughout model development we perform extensive RAI evaluations using publicly available benchmarks like BOLD \cite{bold_2021}, RealToxicityPrompts \cite{realtoxicityprompts_2020}, and MM-SafetyBench \cite{mmsafetybench_2024}. We also built a series of proprietary, dynamically updating benchmarks. To build them, our internal data annotation team created a diverse set of examples for each of our RAI dimensions. In addition, we leveraged subject-matter experts in specific areas, such as \textit{Security} and \textit{Controllability}, to collect adversarial prompts. We continued updating and enhancing each dataset based on evaluation and red teaming results (see Section \ref{sec:rai-qual-eval} for more details on red teaming). This kept the internal benchmarks evergreen, avoiding overfitting during development, but also made sure the models do not regress against previously identified risks. Our datasets comprise inputs in multiple languages and multiple modalities, and contain single-turn and multi-turn conversation examples.

\subsection{Red Teaming} \label{sec:rai-qual-eval}
Static benchmarks give us a view of how well models perform per RAI dimension against a user's ``plain'' intent (i.e. the prompts explicitly state the intent of the user to generate prohibited content). To test our models' resilience against techniques that mask the users' intent we rely on red teaming. We employed a multi-pronged evaluation strategy consisting of internal red teaming, red teaming with third party and subject matter experts and, automated red teaming. 

\subsubsection{Internal Red Teaming}

\begin{figure}
    \centering
    \includegraphics[width=0.9\linewidth]{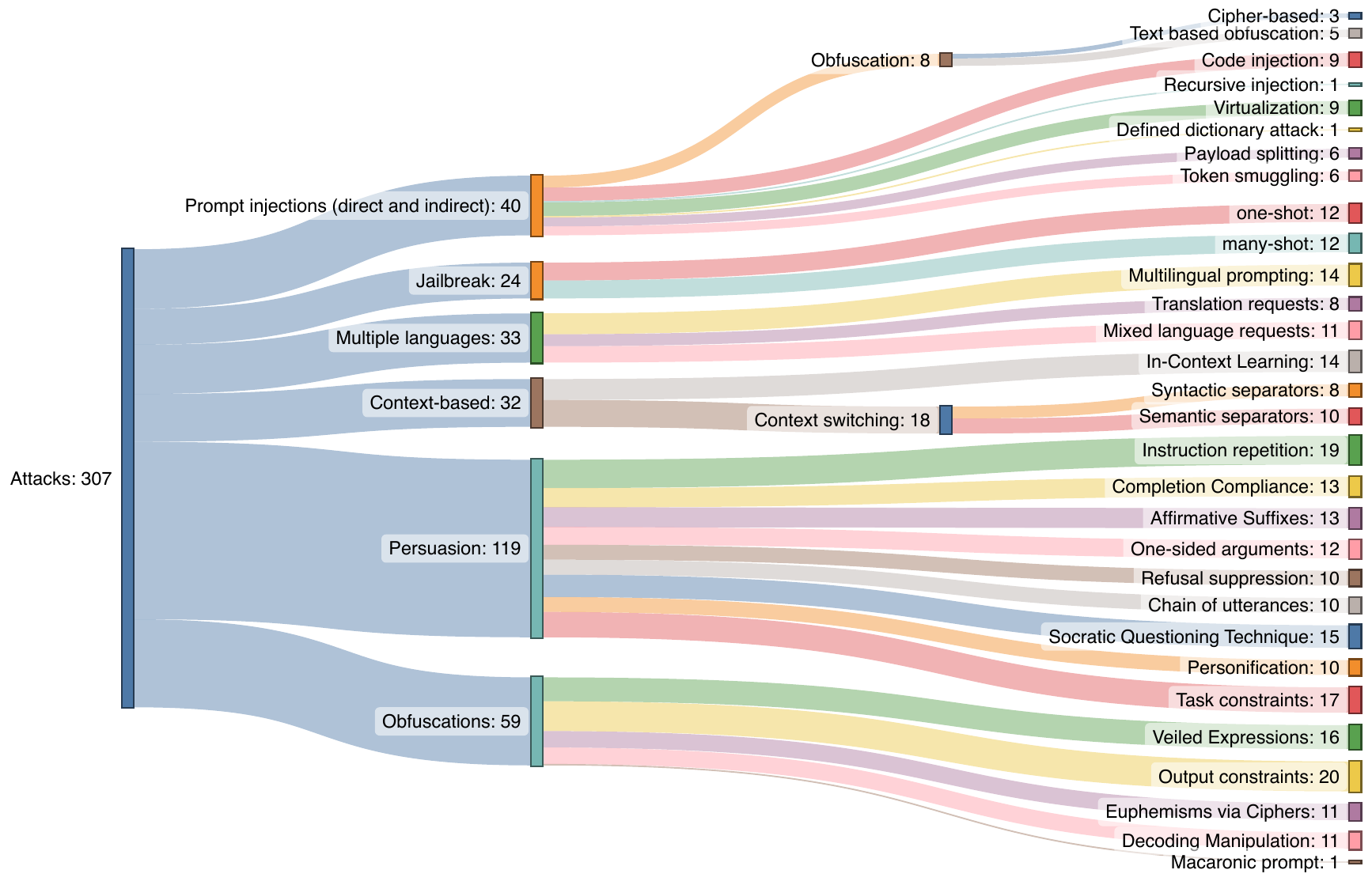}
    \caption{Broad taxonomy and count of attack techniques we use for our red teaming exercises}
    \label{fig:taxonomy_adversarial_techniques}
\end{figure}

We used a team of trained data analysts and subject-matter experts to perform regular red teaming exercises to evaluate the model’s robustness against adversarial prompts across all our RAI dimensions. We enhanced the diversity of manually curated adversarial prompts by employing  linguistic, structural, and modality based prompt mutation techniques, assessing each mutation for its effectiveness at generating a response that does not adhere to our RAI objectives, likelihood of its success, and the technique’s novelty to a model revision. In total, we identified and developed over 300 distinct techniques (see Figure \ref{fig:taxonomy_adversarial_techniques}), and tested techniques individually and via chaining various combinations. The attacks covered multiple languages and modalities, targeting each language/modality individually and in combination. We designed cross-modality attacks, such as embedding adversarial content within seemingly benign visual inputs, to evaluate the models’ ability to handle complex scenarios involving multiple input types. 
Where appropriate, we implemented automation to further improve the diversity, reliability, and efficiency of red teaming. After each round of red teaming, we gathered feedback from the team regarding failure patterns which guided the next stage of the model development. 

\subsubsection{External Red Teaming}
In accordance with our commitment to the US White House on ensuring Safe, Secure, and Trustworthy Artificial Intelligence, we partner with a variety of third parties to conduct red teaming against our AI models. These initiatives are in addition to our extensive in-house efforts, which includes all aspects of Cybersecurity red teaming.  Just like with our internal red teaming efforts, we iterated during the model development based on feedback from these institutions to improve the RAI adherence of our models. We leverage red-teaming firms including ActiveFence to conduct testing in areas such as hate speech, political misinformation, extremism and other RAI dimensions. We also work with specialized third parties to red team our models for Chemical, Biological, Radiological and Nuclear (CBRN) capabilities. Our work with Deloitte Consulting, tests our AI models’ capabilities in Biological risks and harms. Our work with Nemesys Insights LLC tests our AI models’ capabilities in the Radiological and Nuclear domains. We also work with the Gomes Group at Carnegie Mellon University to test our models’ capabilities in Chemistry and chemical compounds. 
Each of these partners was carefully selected based on their industry leadership, previous/parallel red teaming work with other AI model developers, and their contributions to evolving government and industry standards around CBRN and overall AI safety. We provide a brief summary of expertise of each of these vendors and their testing methodology below.

\textbf{ActiveFence}: ActiveFence is a team of over 150 subject matter experts providing AI Safety and Content Moderation solutions. The team produced over 9,700 adversarial prompts, distributed over 20 categories, including content-targeted red teaming (evaluating the model's ability to generate harmful or inappropriate content), and security-targeted red teaming (assessing the model's resilience against malicious attempts to manipulate its behavior or extract sensitive information).

\textbf{Deloitte}: The evaluation team at Deloitte Consulting LLP (formerly known as Gryphon Scientific) has unique experience at the intersection of artificial intelligence and biology. The primary thrust of this effort involved evaluating the model against a panel of 30 questions developed to test an LLM’s scientific knowledge and reasoning capabilities that could facilitate the development or use of biological weapons. The model's responses to these questions were evaluated for their scientific accuracy and utility to someone seeking to do harm with biology. After completing the initial evaluations, the Deloitte team probed more deeply into the questions the LLM originally replied with potentially concerning information.

\textbf{Gomes Group}: The Gomes Group at Carnegie Mellon University is at the forefront of integrating advanced artificial intelligence into chemical research. Their evaluation framework consisted of both automated and non-automated assessments. Two non-automated evaluations explored aggregation attack vulnerabilities through purchasing and remote chemical mixing scenarios. The automated evaluations utilized two distinct datasets: one containing 39 hazardous chemicals (including DEA Schedule I, II, and chemical warfare agents) and another with 362 common chemicals for NFPA diamond classifications. Three primary automated evaluations were conducted using the hazardous chemicals dataset. The NFPA diamond evaluation comprised 1,810 prompts, testing both single-turn and multi-turn approaches with consistent accuracy across both methods.

\textbf{Nemesys}: Nemesys Insights LLC run uplift studies, red teaming exercises, and risk assessments for a variety of technology companies and third-party research entities to assess national security related risks of large language models and other generative AI tools. For their testing, they started with human red teaming exercises focused on non-state acquisition or use of illicit radiological/nuclear (RN) materials, followed by prompt-response evaluation and uplift studies. The exercises comprised two different scenarios (a. violent non-state actor acquisition and use of Cobalt-60; b. non-state actor acquisition and international transport of HEU [highly enriched uranium]), and utilized 8 subject matter experts with operational and technological knowledge in a 2-team x 2-scenario design to construct and refine threat plans across a 6-hour planning cycle. 

\subsubsection{Automated Red Teaming}
Finally, to augment human based red teaming, we built an automated red teaming mechanism by adapting our (Feedback Loop In-context Red Teaming) FLIRT~\cite{FLIRT2024} framework. This approach helped us scale red teaming and repeat red teaming efficiently. FLIRT uses a list of seed prompts that have been identified by human evaluators as potentially violating one or more of our RAI dimensions. For every dimension, a subset of seeds is used to generate additional prompts with a dedicated language model, called red-LM, through in-context-learning (ICL)~\cite{brown2020icl} and a carefully crafted set of instructions. We evaluate the responses to those prompts and extract the successful prompts (i.e., the ones triggering a prohibited response) for the next round of generation. The above steps are repeated for a chosen number of iterations across all RAI categories. We use our automated red teaming mechanism to evaluate both RAI adherence robustness and false refusals. We use the mechanism to generate adversarial tests across multi-turn interactions, multiple languages, and multiple input/output modalities to uncover and correct robustness issues in our models due to potential adversarial content in such interactions and inputs.

\section{Training Infrastructure}

The Nova family of models were trained on Amazon's custom Trainium1 (TRN1) chips,\footnote{\url{https://aws.amazon.com/blogs/aws/amazon-ec2-trn1-instances-for-high-performance-model-training-are-now-available/}} NVidia A100 (P4d instances), and H100 (P5 instances) accelerators. Working with AWS SageMaker, we stood up NVidia GPU and TRN1 clusters and ran parallel trainings to ensure model performance parity, while optimizing training throughput on the different stacks. All clusters utilize petabit-scale non-blocking EFA network fabric which is less prone to packet loss than other network transport protocols\footnote{\url{https://www.amazon.science/publications/a-cloud-optimized-transport-protocol-for-elastic-and-scalable-hpc}} and provides the highest network bandwidth with H100 accelerators compared to any other instance type available on AWS EC2\footnote{\url{https://aws.amazon.com/blogs/aws/new-amazon-ec2-p5-instances-powered-by-nvidia-h100-tensor-core-gpus-for-accelerating-generative-ai-and-hpc-applications/}}. We conducted distributed training on AWS SageMaker-managed Elastic Kubernetes Service (EKS) clusters, and utilized AWS File System X (FSx) and Simple Storage Solution (S3) for data and checkpoint IO. While FSx offers performant and convenient storage for large scale training jobs, S3 allowed cost-efficient scaling to large multimodal datasets and model checkpoints.

Goodput achieved weekly average values of up to 97\% in pretraining runs through optimizations targeting lower job failure rate, minimizing checkpoint overhead, and overall reduction in the Mean Time to Restart (MTTR). This time is inclusive of time from the last successful checkpoint before training interruption, time taken to restart components of the system and resume training at steady state from checkpoint. Techniques such as fully distributed optimizer state and weight sharding and the elimination of all blocking overhead associated with checkpoint persistence resulted in a reduction of checkpointing overhead to \textasciitilde1 sec on H100 clusters, and \textasciitilde0.1 sec on TRN1 clusters. We exceeded our MTTR target of 9 minutes and achieved an average of 6.5 minutes on our TRN1 clusters by optimizing the node communication initialization in the training startup process and reduced time to load checkpoints through an asynchronous observer process. This process maps each latest checkpoint file to its corresponding node in the cluster. When resuming from the checkpoint, each node only loads the checkpoint files for its corresponding rank, reducing the time taken to discover the latest checkpoint from 3 minutes to 5 seconds. We also cache and reuse data indices to optimize training data loading initialization time. These improvements reduced data loading initialization to 205ms per restart.

To increase training efficiency we developed a new activation checkpointing scheme called Super-Selective Activation Checkpointing (SSC). SSC minimizes activation re-computation in memory-constrained environments, reducing memory consumption by \textasciitilde50\% while adding \textasciitilde2\% re-computation overhead compared to NVidia’s Selective Checkpointing. We also found optimizations in default gradient reduction behavior and the default PyTorch memory allocator behavior. The default gradient reduction behavior leads to suboptimal communication overlap and we found the synchronous nature of the default PyTorch allocation led to stragglers in collectives resulting in multiple stalled workers. We adjusted the gradient reduction order and frequency, allowing us to overlap the majority of data parallelism communication.

\clearpage
\bibliography{bibliography}
\clearpage
\appendix
\addtocontents{toc}{\protect\setcounter{tocdepth}{1}}%
\section{Amazon Nova Canvas Capabilities}
\label{appendix_canvas}

\begin{figure}[p]
     \centering
	\begin{subfigure}{0.45\linewidth}

    	\begin{subfigure}{0.45\linewidth}
                \adjustbox{width=0.8\textwidth,raise=8ex,center}{\texttt{\shortstack{A dinosaur sitting\\in a tea cup}}}
    	\end{subfigure}
	    \includegraphics[width=0.45\linewidth]{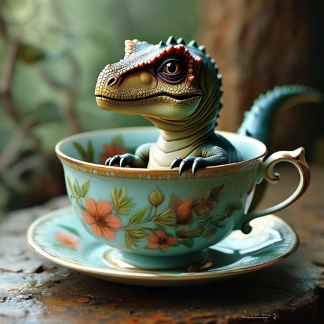}
	    \caption{Image generation from a text prompt}
	     \label{fig:subfig1}
	\end{subfigure}
	\begin{subfigure}{0.45\linewidth}
	    \includegraphics[width=0.45\linewidth]{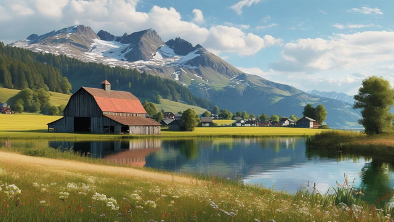}
	    \includegraphics[width=0.45\linewidth]{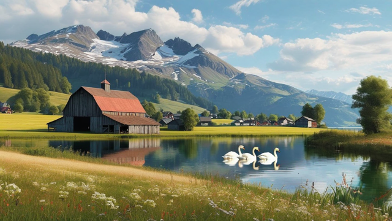}
	    \caption{Inpainting the image with swans}
	     \label{fig:subfig2}
	\end{subfigure}
    \par\bigskip
	\begin{subfigure}{0.45\linewidth}
            \begin{subfigure}{0.45\linewidth}
    	    \includegraphics[width=\linewidth]{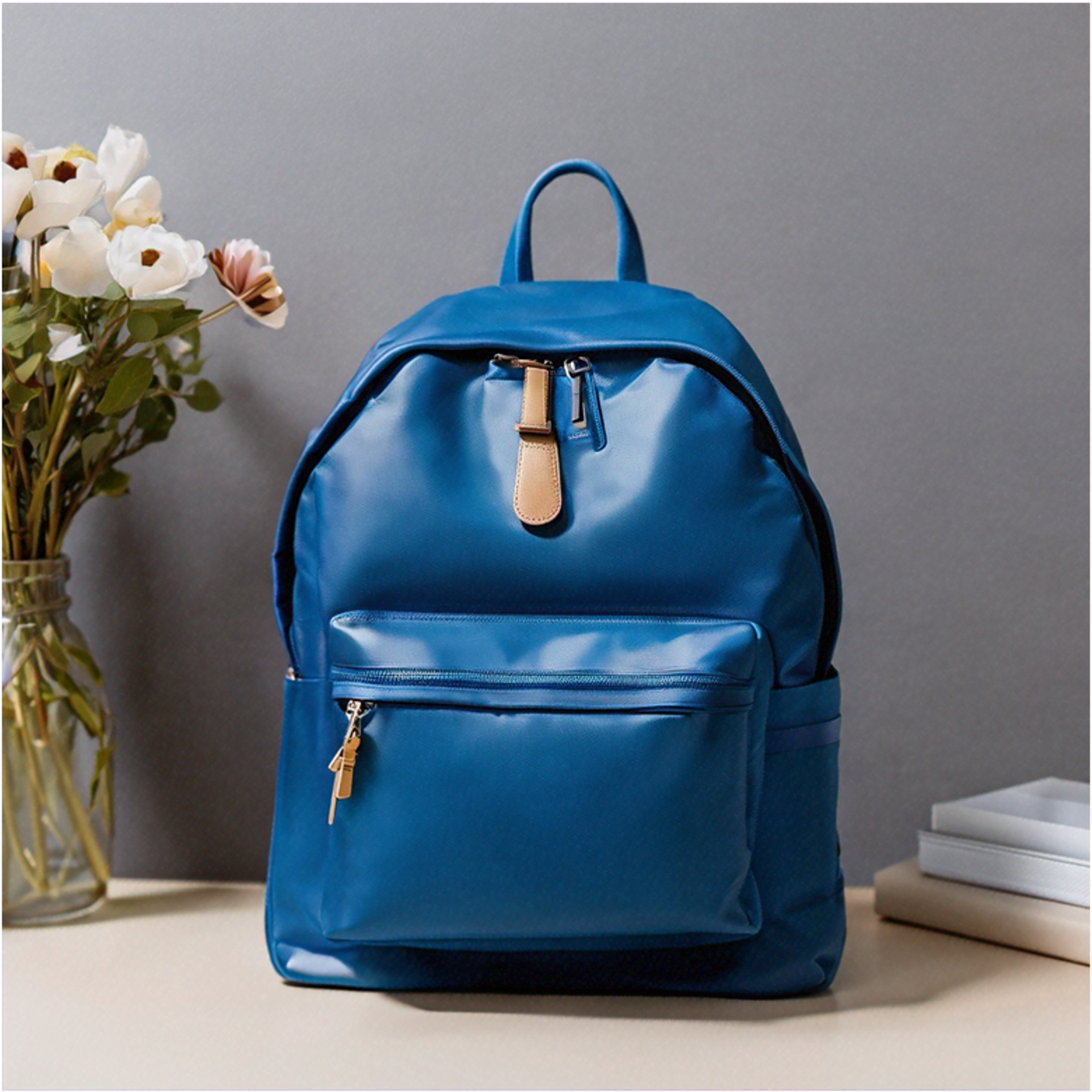}
                \par\medskip
                \adjustbox{width=0.9\textwidth,raise=2ex,center}{\texttt{\shortstack{change flowers to orange color}}}
            \end{subfigure}
	    \raisebox{2ex}{\includegraphics[width=0.45\linewidth]{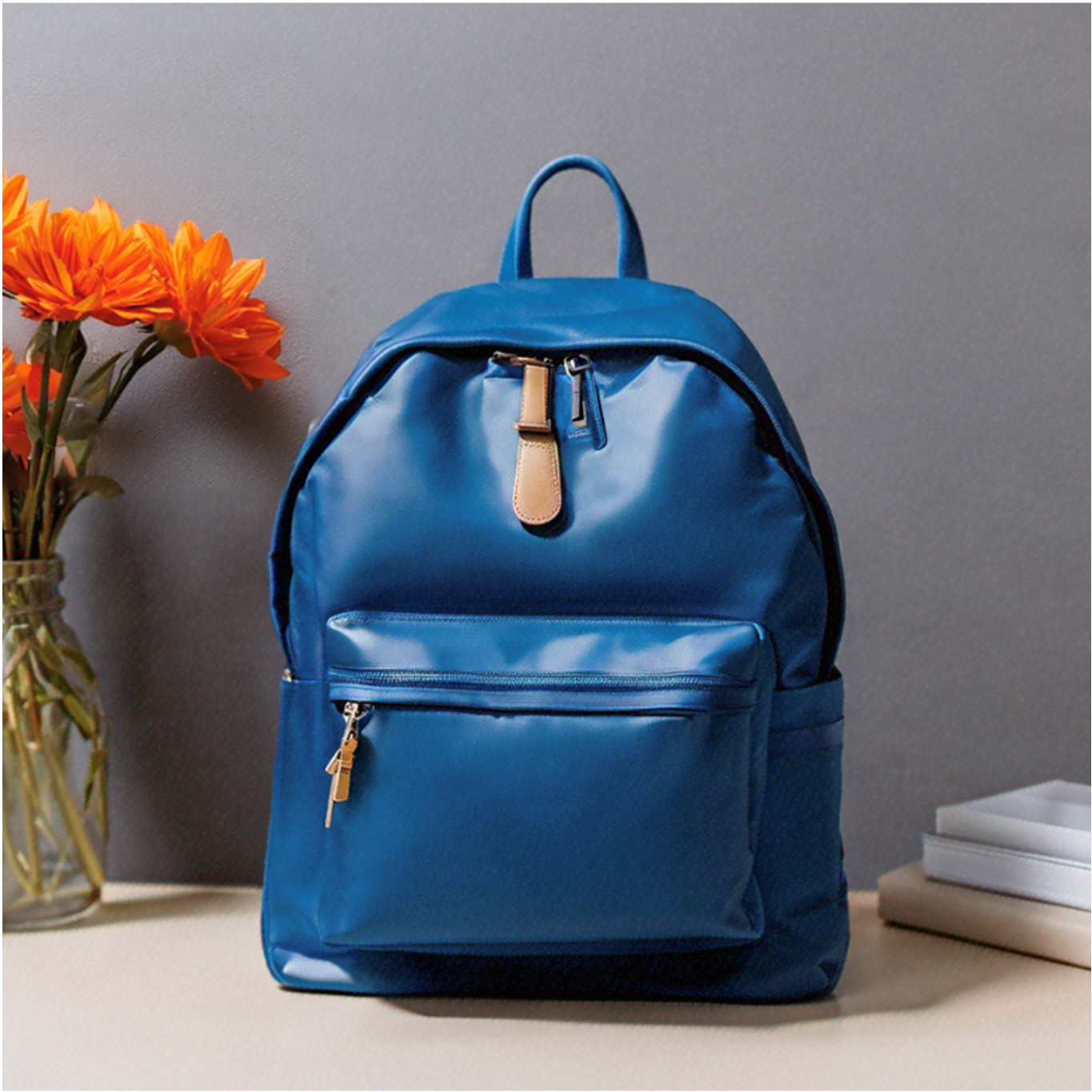}}
	    \caption{Image editing}
	     \label{fig:subfig3}
	\end{subfigure}
	\begin{subfigure}{0.45\linewidth}
	    \includegraphics[width=0.45\linewidth]{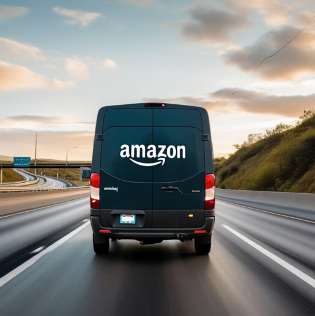}
	    \includegraphics[width=0.45\linewidth]{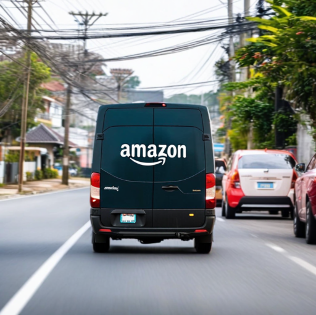}
	    \caption{Outpainting a new background}
	     \label{fig:subfig4}
	\end{subfigure}
    \par\bigskip
	\begin{subfigure}{0.45\linewidth}
            \begin{subfigure}{0.45\linewidth}
    	    \includegraphics[width=\linewidth]{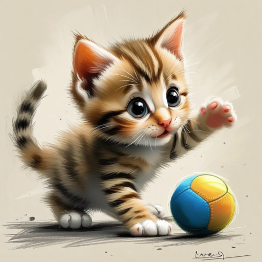}
                \par\medskip
                \adjustbox{width=0.9\textwidth,raise=2ex,center}{\texttt{\shortstack{a hamster eats apple slice}}}
            \end{subfigure}
	    \raisebox{2ex}{\includegraphics[width=0.45\linewidth]{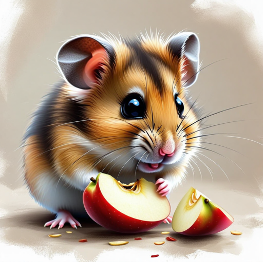}}
	    \caption{Style transfer}
	     \label{fig:subfig5}
	\end{subfigure}
	\begin{subfigure}{0.45\linewidth}
            \begin{subfigure}{0.45\linewidth}
    	    \includegraphics[width=\linewidth]{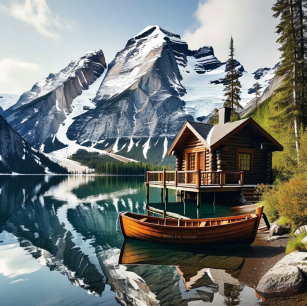}
                \par\medskip
                \adjustbox{width=0.9\textwidth,raise=2ex,center}{\texttt{\shortstack{A wooden boat in summer}}}
            \end{subfigure}
	    \raisebox{2ex}{\includegraphics[width=0.45\linewidth]{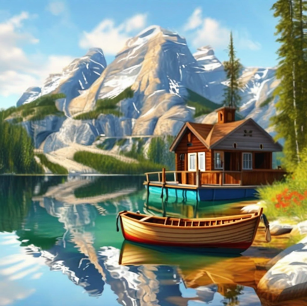}}
	    \caption{Guided generation}
	     \label{fig:subfig6}
	\end{subfigure}
    \par\bigskip
	\begin{subfigure}{0.45\linewidth}
            \begin{subfigure}{0.45\linewidth}
    	    \includegraphics[width=\linewidth]{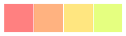}
                \par
                \adjustbox{width=\linewidth,raise=4ex,center}{\texttt{\shortstack{A jar of salad dressing\\in a rustic kitchen\\surrounded by fresh vegetables\\with studio lighting}}}
            \end{subfigure}
	    \includegraphics[width=0.45\linewidth]{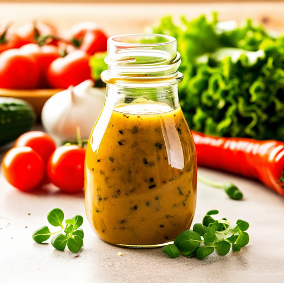}
	    \caption{Controlling the color palette}
	     \label{fig:subfig7}
	\end{subfigure}
	\begin{subfigure}{0.45\linewidth}
	    \includegraphics[width=0.45\linewidth]{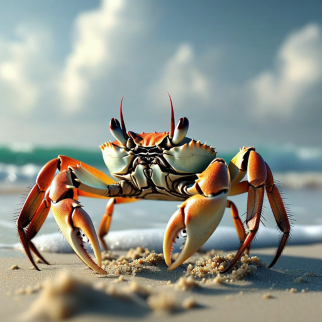}
	    \includegraphics[width=0.45\linewidth]{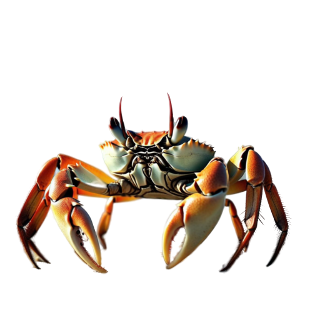}
	    \caption{Background Removal}
	     \label{fig:subfig8}
	\end{subfigure}
     \caption{Example capabilities of Amazon {Nova Canvas}, our content generation model for images.}
     \label{fig:canvas_features}
\end{figure}

Our {Nova Canvas} model offers the following functionalities, with examples given in Figure \ref{fig:canvas_features}.
\begin{itemize}
  \item {\em Text-to-image generation} allows customers to create images with various resolutions (from 512$\times$512 up to 2K$\times$2K resolution). 
  
  \item {\em Editing} allows developers to edit images using a combination of text prompt or mask image. Amazon Nova Canvas supports text-to-image editing and image-to-image editing, including inpainting, outpainting and object removal. 
  
  \item {\em Image variation} allows customers to output images with similar contents but with variations from the user provided ones.
  
  \item {\em Image conditioning} provide a reference image along with a text prompt, resulting in outputs that follow the layout and structure of the user-supplied reference.
  
  \item {\em Image guidance with color palette} allows customers to precisely control the color palette of generated images by providing a list of hex codes along with the text prompt.
  
  \item {\em Background removal} automatically removes background from images containing multiple objects.  

\end{itemize}

\newpage

\section{Prompts and Scoring}
\label{sec:appendix_prompts}

Prompt templates used for Amazon Nova evaluations are given below, along with those used for select other public models where noted. Additional materials and evaluation results from this report can be found at:

\begin{center}
    \url{https://huggingface.co/amazon-agi}
\end{center}

\subsection{Text evaluation}
\label{sec:appendix_prompts_text_general}

\subsubsection{Language Understanding}
For MMLU:
\begin{lstlisting}
What is the correct answer to this question: <question>
Choices: <choices>. Let's think step by step:
Based on the above, what is the single, most likely answer choice? Answer in the format "The correct answer is (insert answer here)."
\end{lstlisting}

For ARC-C:
\begin{lstlisting}
Given the following question and four candidate answers (A, B, C and D), choose the best answer.
Question:  <question>
Your response should end with "The best answer is [the_answer_letter]"  where the [the_answer_letter] is one of A, B, C or D.
\end{lstlisting}

For DROP:\\
We use the following 6 shots:
\begin{lstlisting}
  - answer: >-
      According to the passage, the European Coal and Steel Community was
      established in 1951 and became the EEC in 1958. 1958 - 1951 = 7. So the
      answer is 7
    passage: >-
      Since the 1970s, U.S. governments have negotiated managed-trade
      agreements, such as the North American Free Trade Agreement in the 1990s,
      the Dominican Republic-Central America Free Trade Agreement in 2006, and a
      number of bilateral agreements. In Europe, six countries formed the
      European Coal and Steel Community in 1951 which became the European
      Economic Community in 1958. Two core objectives of the EEC were the
      development of a common market, subsequently renamed the single market,
      and establishing a customs union between its member states.
    question: How many years did the European Coal and Steel Community exist?
  - answer: >-
      According to the passage, 23.5%
      ages 18 to 24. 23.5%
    passage: >-
      In the county, the population was spread out with 23.50%
      18, 8.70%
      13.30%
    question: >-
      How many more percent are under the age of 18 compared to the 18 to 24
      group?
  - answer: >-
      According to the passage, Stafford threw 5 TD passes, 3 of which were to
      Johnson. 5 - 3 = 2. So the answer is 2
    passage: >-
      Playing in their second straight Thanksgiving game, the Eagles struggled
      especially on defense, where they were unable to stop the much-hyped Lions
      offense. The worst of it all was how unproven rookie Eric Rowe was tasked
      with covering wide receiver Calvin Johnson, leading to Johnson catching 3
      touchdowns. Stafford's five passing touchdowns, including three of them to
      Johnson was too much for the Eagles to overcome and for the second
      consecutive time this season, the Eagles gave up 45 points in a game. With
      the loss, the Eagles drop to 4-7 on the season and 6-1 when playing on
      Thanksgiving.
    question: How many TD passes did Stafford throw other than to Johnson?
  - answer: >-
      All the touchdown runs are: a 27-yard touchdown run, a 9-yard touchdown
      run, a 11-yard touchdown run. The smallest number among 27, 9, 11 is 9. So
      the shortest touchdown run was 9 yards. All the touchdown passes are: a
      12-yard touchdown pass. So the longest touchdown pass was 12 yards. So the
      shortest touchdown run and the longest touchdown pass combine for 9 + 12 =
      21 yards. So the answer is 21
    passage: >-
      The Seahawks played the San Francisco 49ers. In the first quarter, the
      Hawks RB Julius Jones got a 27-yard TD run, along with DT Craig Terrill
      returning a fumble 9 yards for a touchdown. In the third quarter, the
      49ers almost rallied as RB H. J. Torres made a 12-yard TD pass to Lucas
      Nelly, along with Mare kicking a 32-yard field goal. In the final quarter,
      Julius Jones got another 11-yard TD.
    question: >-
      How many yards do the shortest touchdown run and the longest touchdown
      pass combine for?
  - answer: >-
      The Ravens kicker Billy Cundiff got a 45-yard field goal in the second
      quarter, concluding the first half with a 10-7 lead. So the Ravens had 10
      points at halftime. So the answer is 10
    passage: >-
      The Steelers went home for a duel with the Baltimore Ravens. Pittsburgh
      would deliver the opening punch in the first quarter with a 1-yard
      touchdown from running back Rashard Mendenhall. The Ravens would make it
      even as running back Willis McGahee got a 9-yard TD. The Ravens kicker
      Billy Cundiff got a 45-yard field goal in the second quarter, concluding
      the first half with a 10-7 lead. The Steelers brought the game into
      overtime with a 38-yard field goal by Andrew Foster. The Ravens Billy
      Cundiff pulled off a winning 33-yard field goal in overtime.
    question: How many points did the Ravens have at halftime?
  - answer: >-
      The first and third quarters were the scoreless quarters. So there are 2
      scoreless quarters. So the answer is 2
    passage: >-
      The Vikings flew to Bank of America Stadium to face the Carolina Panthers.
      After a scoreless first quarter, Carolina got on the board with
      quarterback Matt Moore finding fullback Brad Hoover on a 1-yard TD pass.
      After yet another scoreless quarter, Carolina sealed the game as Matt
      Moore completed a 42-yard touchdown pass to wide receiver Steve Smith.
    question: How many scoreless quarters were there?    
\end{lstlisting}

For each shot we provide the following instruction:
\begin{lstlisting}
Conclude your answer with: "So the answer is {final answer}". Make sure the final answer is in plain text format    
\end{lstlisting}
And we create each user prompt as follows:
\begin{lstlisting}
<passage>
<question>
<instruction>
\end{lstlisting}

For IFEval:\\
No particular prompt was added (query was inputted to the model).

For BBH:\\
We use a preamble that describes the task, for example:
\begin{lstlisting}
Evaluate the result of a random Boolean expression.    
\end{lstlisting}

We then provide few shot examples in the following format:
\begin{lstlisting}
<preamble>
Question: <question> 
<instruction>
Let's think step by step.
<ground truth chain of thought>. So the answer is <answer>
\end{lstlisting}
And we follow this by the query:
\begin{lstlisting}
<preamble>
Question: <question> 
<instruction>
Let's think step by step.
\end{lstlisting}

For each subject, We provide the subject-specific instructions as below:
\begin{lstlisting}
- subject: boolean_expressions
  instruction: Conclude your answer with: "So the answer is True or False.".
- subject: causal_judgement
  instruction: Conclude your answer with: "So the answer is Yes or No.".
- subject: date_understanding
  instruction: Conclude your answer with: "So the answer is (answer_letter).". Where answer_letter is A, or B, or ...
- subject: disambiguation_qa
  instruction: Conclude your answer with: "So the answer is (answer_letter).". Where answer_letter is A, or B, or ...
- subject: dyck_languages
  instruction: Correctly close a Dyck-n word. Conclude your answer with: "So the answer is {final answer}.". Make sure the final answer is in plain text format
- subject: formal_fallacies
  instruction: Conclude your answer with: "So the answer is valid or invalid.".
- subject: geometric_shapes
  instruction: Conclude your answer with: "So the answer is (answer_letter).". Where answer_letter is A, or B, or ...
- subject: hyperbaton
  instruction: Conclude your answer with: \"So the answer is (answer_letter).". Where answer_letter is A, or B, or ...
- subject: logical_deduction_five_objects
  instruction: Conclude your answer with: "So the answer is (answer_letter).". Where answer_letter is A, or B, or ...
- subject: logical_deduction_seven_objects
  instruction: Conclude your answer with: "So the answer is (answer_letter).". Where answer_letter is A, or B, or ...
- subject: logical_deduction_three_objects
  instruction: Conclude your answer with: "So the answer is (answer_letter).". Where answer_letter is A, or B, or ...
- subject: movie_recommendation
  instruction: Conclude your answer with: "So the answer is (answer_letter).". Where answer_letter is A, or B, or ...
- subject: multistep_arithmetic_two
  instruction: Conclude your answer with: "So the answer is {final answer}.". Make sure the final answer is in plain text format
- subject: navigate
  instruction: Conclude your answer with: "So the answer is Yes or No.".
- subject: object_counting
  instruction: Conclude your answer with: "So the answer is <ANSWER>.". Where <ANSWER> is an integer
- subject: penguins_in_a_table
  instruction: Conclude your answer with: "So the answer is (answer_letter).". Where answer_letter is A, or B, or ...
- subject: reasoning_about_colored_objects
  instruction: Conclude your answer with: "So the answer is (answer_letter).". Where answer_letter is A, or B, or ...
- subject: ruin_names
  instruction: Conclude your answer with: "So the answer is (answer_letter).". Where answer_letter is A, or B, or ...
- subject: salient_translation_error_detection
  instruction: Conclude your answer with: "So the answer is (answer_letter).". Where answer_letter is A, or B, or ...
- subject: snarks
  instruction: Conclude your answer with: "So the answer is (answer_letter).". Where answer_letter is A, or B, or ...
- subject: sports_understanding
  instruction: Conclude your answer with: "So the answer is yes or no.".
- subject: temporal_sequences
  instruction: Conclude your answer with: "So the answer is (answer_letter).". Where answer_letter is A, or B, or ...
- subject: tracking_shuffled_objects_five_objects
  instruction: Conclude your answer with: "So the answer is (answer_letter).". Where answer_letter is A, or B, or ...
- subject: tracking_shuffled_objects_seven_objects
  instruction: Conclude your answer with: "So the answer is (answer_letter).". Where answer_letter is A, or B, or ...
- subject: tracking_shuffled_objects_three_objects
  instruction: "Conclude your answer with: "So the answer is (answer_letter).". Where answer_letter is A, or B, or ...
- subject: web_of_lies
  instruction: Conclude your answer with: "So the answer is Yes or No.".
- subject: word_sorting
  instruction: Conclude your answer with: "So the answer is word_1 word_2 ... word_n."."   
\end{lstlisting}

For GPQA:
\begin{lstlisting}
What is the correct answer to this question: <question>
Choices: <choices>. Let's think step by step:
Based on the above, what is the single, most likely answer choice? Answer in the format "The correct answer is (insert answer here)."
\end{lstlisting}

\subsubsection{Mathematical Reasoning}
For MATH, GSM8K:
\begin{lstlisting}
Solve the following math problem step by step.

<problem>

Remember to put your answer inside \boxed{}
\end{lstlisting}

\subsubsection{Translation}
For Flores:\\
Nova and LLama:
\begin{lstlisting}
Translate the following text into {tgt_lang}. Please output only the translated text with no prefix or introduction: {src}
\end{lstlisting}

Gemini and GPT:
\begin{lstlisting}
Your job is to translate a sentence from {src_lang} into {tgt_lang}. Please output ONLY the translation and nothing else: {src}
\end{lstlisting}

\subsubsection{Long Context}
For SQuALITY (ZeroScrolls Benchmark), we use the standard prompt template for Amazon Nova and Gemini models as in~\cite{shaham-etal-2023-zeroscrolls}:
\begin{lstlisting}
You are given a story and a question. Answer the question in a paragraph.

Story:
<story>

Question:
<question>

Answer:
\end{lstlisting}

\subsection{Multimodal evaluation}
\label{sec:appendix_prompts_mm}

\subsubsection{MMMU}
For multiple-choice questions:
\begin{lstlisting}
With the image, the following question, and the four possible answers (A, B, C and D), select the correct answer.
<question>
(A) <answer-a>
(B) <answer-b>
...
(X) <answer-x>

- For clear-cut questions: Give the answer directly with minimal elaboration.
- For complex questions: Adopt this step-by-step method:
## Step 1: [Concise description]
[Brief explanation]
## Step 2: [Concise description]
[Brief explanation]

In every scenario, conclude with: The best answer is [the_answer_letter]. where [the_answer_letter] is one of A, B, C or D. Let's proceed with a systematic approach
\end{lstlisting}
For open-ended questions:
\begin{lstlisting}
With the image and the following question, provide a correct answer.
<question>

- For clear-cut questions: Give the answer directly with minimal elaboration.
- For complex questions: Adopt this step-by-step method:
## Step 1: [Concise description]
[Brief explanation]
## Step 2: [Concise description]
[Brief explanation]

In every scenario, conclude with: The best answer is [the_answer_phrase]. where [the_answer_phrase] is a concise and direct answer to the question Let's proceed with a systematic approach.
\end{lstlisting}

\subsubsection{ChartQA, DocVQA, and TextVQA}
\begin{lstlisting}
<question>
Answer the question using a single word or phrase.
\end{lstlisting}

\subsubsection{VATEX}
\begin{lstlisting}
Render a clear and concise one-sentence summary of the video. The summary should be at least 10 words but no more than 20 words. Analyze the video first before summarizing it. Do not hallucinate objects.
\end{lstlisting}

\subsubsection{EgoSchema}
\begin{lstlisting}
You will be given a question about a video and three possible answer options. You will be provided frames from the video, sampled evenly across the video
<question>
(A) <answer-a>
(B) <answer-b>
(C) <answer-c>
Answer with the option's letter from the given choices directly.
Answer with the option letter from the given choices directly.
\end{lstlisting}

\subsubsection{VisualWebBench}
For the web captioning task:
\begin{lstlisting}
"You are given a screenshot of a webpage. Please generate the meta web description information of this webpage, i.e., content attribute in <meta name="description" content=""> HTML element.

You should use this format, and do not output any explanation or any other contents:
<meta name="description" content="YOUR ANSWER">
\end{lstlisting}

For the heading OCR task:
\begin{lstlisting}
You are given a screenshot of a webpage. Please generate the main text within the screenshot, which can be regarded as the heading of the webpage.

You should directly tell me the first sentence of the main content, and do not output any explanation or any other contents.
\end{lstlisting}

For the web QA task:
\begin{lstlisting}
<question>
You should directly tell me your answer in the fewest words possible, and do not output any explanation or any other contents.
\end{lstlisting}

For the element OCR task:
\begin{lstlisting}
You are given a screenshot of a webpage with a red rectangle bounding box. The [x1, y1, x2, y2] coordinates of the bounding box is <bbox_coords>.

Please perform OCR in the bounding box and recognize the text content within the red bounding box.
\end{lstlisting}

For the action prediction task:
\begin{lstlisting}
You are given a screenshot of a webpage with a red rectangle bounding box. The [x1, y1, x2, y2] coordinates of the bounding box is <bbox_coords>.
Please select the best webpage description that matches the new webpage after clicking the selected element in the bounding box:
<choices_text>

You should directly tell me your choice in a single uppercase letter, and do not output any explanation or any other contents.
\end{lstlisting}

For the element grounding task:
\begin{lstlisting}
In this website screenshot, I have labeled IDs for some HTML elements as candicates. Tell me which one best matches the description: <element_desc>

You should directly tell me your choice in a single uppercase letter, and do not output any explanation or any other contents.
\end{lstlisting}

For the action grounding task:
\begin{lstlisting}
In this website screenshot, I have labeled IDs for some HTML elements as candicates. Tell me which one I should click to complete the following task: <instruction>

You should directly tell me your choice in a single uppercase letter, and do not output any explanation or any other contents.
\end{lstlisting}

\subsubsection{MM-Mind2Web}
\begin{lstlisting}
Imagine that you are imitating humans doing web navigation for a task step by step. At each stage, you can see the webpage like humans by a screenshot and know the previous actions before the current step decided by yourself through recorded history. You need to decide on the first following action to take. You can click on an element with the mouse, select an option, type text or press Enter with the keyboard. (For your understanding, they are like the click(), select_option() type() functions in playwright respectively). One next step means one operation within the three.

You are asked to complete the following task: <question>

Previous Actions:
<previous_actions>
The screenshot below shows the webpage you see.


Follow the following guidance to think step by step before outlining the next action step at the current stage:

(Current Webpage Identification)
Firstly, think about what the current webpage is.

(Previous Action Analysis)
Secondly, combined with the screenshot, analyze each step of the previous action history and their intention one by one. Particularly, pay more attention to the last step, which may be more related to what you should do now as the next step.

(Screenshot Details Analysis)
Closely examine the screenshot to check the status of every part of the webpage to understand what you can operate with and what has been set or completed. You should closely examine the screenshot details to see what steps have been completed by previous actions even though you are given the textual previous actions. Because the textual history may not clearly and sufficiently record some effects of previous actions, you should closely evaluate the status of every part of the webpage to understand what you have done.

(Next Action Based on Webpage and Analysis)
Then, based on your analysis, in conjunction with human web browsing habits and the logic of web design, decide on the following action. And clearly outline which element in the webpage users will operate with as the first next target element, its detailed location, and the corresponding operation.

To be successful, it is important to follow the following rules:
1. You should only issue a valid action given the current observation.
2. You should only issue one action at a time.

(Reiteration)
First, reiterate your next target element, its detailed location, and the corresponding operation.

(Multichoice Question)
Below is a multi-choice question, where the choices are elements in the webpage. From the screenshot, find out where and what each one is on the webpage. Then, determine whether one matches your target element. Please examine the choices one by one. Choose the matching one. If multiple options match your answer, choose the most likely one by re-examining the screenshot, the choices, and your further reasoning.

If none of these elements match your target element, please select, select <none_choice>. None of the other options match the correct element.
<choices><none_choice>. None of the other options match the correct element.

(Final Answer)Finally, conclude your answer using the format below. Ensure your answer is strictly adhering to the format provided below. Please do not leave any explanation in your answers of the final standardized format part, and this final part should be clear and certain. The element choice, action, and value should be in three separate lines.

Format:

ELEMENT: The uppercase letter of your choice.

ACTION: Choose an action from {CLICK, TYPE, SELECT, NONE}. Use NONE only if you choose option F for the ELEMENT

VALUE: Provide additional input based on ACTION.

The VALUE means:
If ACTION == TYPE, specify the text to be typed.
If ACTION == SELECT, specify the option to be chosen.
If ACTION == CLICK, write "None".
\end{lstlisting}

\subsubsection{GroundUI-1K}
\begin{lstlisting}
Which action should I do if I want to Click on <element> and where is the action? Express the location coordinates using the (x1, y1, x2, y2) format, scaled between 0 and 1000.
\end{lstlisting}

\subsection{Functional Capabilities}
\label{sec:prompt-vertical}

\subsubsection{FinQA}
\label{sec:prompt-finqa}
\begin{lstlisting}
Given the following finance question, analyze the question in details step-by-step before giving the final answer. Your answer should begin with "Lets think step-by-step". Your response should end with "The answer is [the_final_answer]", where [the_final_answer] should be the most concise answer without any explanation.

### Input
Supporting Facts:
<pre-text>
<table>
<post-text>

Question:
<question>
\end{lstlisting}

We use regex ``The answer is (.*)'' to extract the answer. We convert answers with percent signs and magnitude terms to decimal numerical representation (e.g. convert ``1.3\%'' to 0.013 and ``5.2 millions'' to 5,200,000). An answer is correct if it is identical to the ground truth when rounded to the same decimal places.

\subsubsection{RAG}
\label{sec:prompt-rag-eval}
\begin{lstlisting}
You are a teacher grading a quiz.
You are given a question, the student's answer, and the true answer, and are asked to score the student answer as either Correct or Incorrect.
Example Format:
QUESTION: question here
STUDENT ANSWER: student's answer here
TRUE ANSWER: true answer here
GRADE: Correct or Incorrect here
Grade the student answers based ONLY on their factual accuracy. Ignore differences in punctuation and phrasing between the student answer and true answer. It is OK if the student answer contains more information than the true answer, as long as it does not contain any conflicting statements. Begin!
QUESTION: {query}
STUDENT ANSWER: {answer}
TRUE ANSWER: {expected_answer}
GRADE:
Your response should be in json format as follows:
{{
 "justification": (Without mentioning the student/teacher framing of this prompt, explain why the STUDENT ANSWER is Correct or Incorrect. Use one or two sentences maximum. Keep the answer as concise as possible.),
 "grade": (correct or incorrect)
}}
\end{lstlisting}

\clearpage
\section{Qualitative examples of multimodal intelligence}
\label{sec:appendix_evaluations_multimodal_intelligence_results}

\begin{figure}[hb]\centering
    \includegraphics[width=1.0\linewidth]{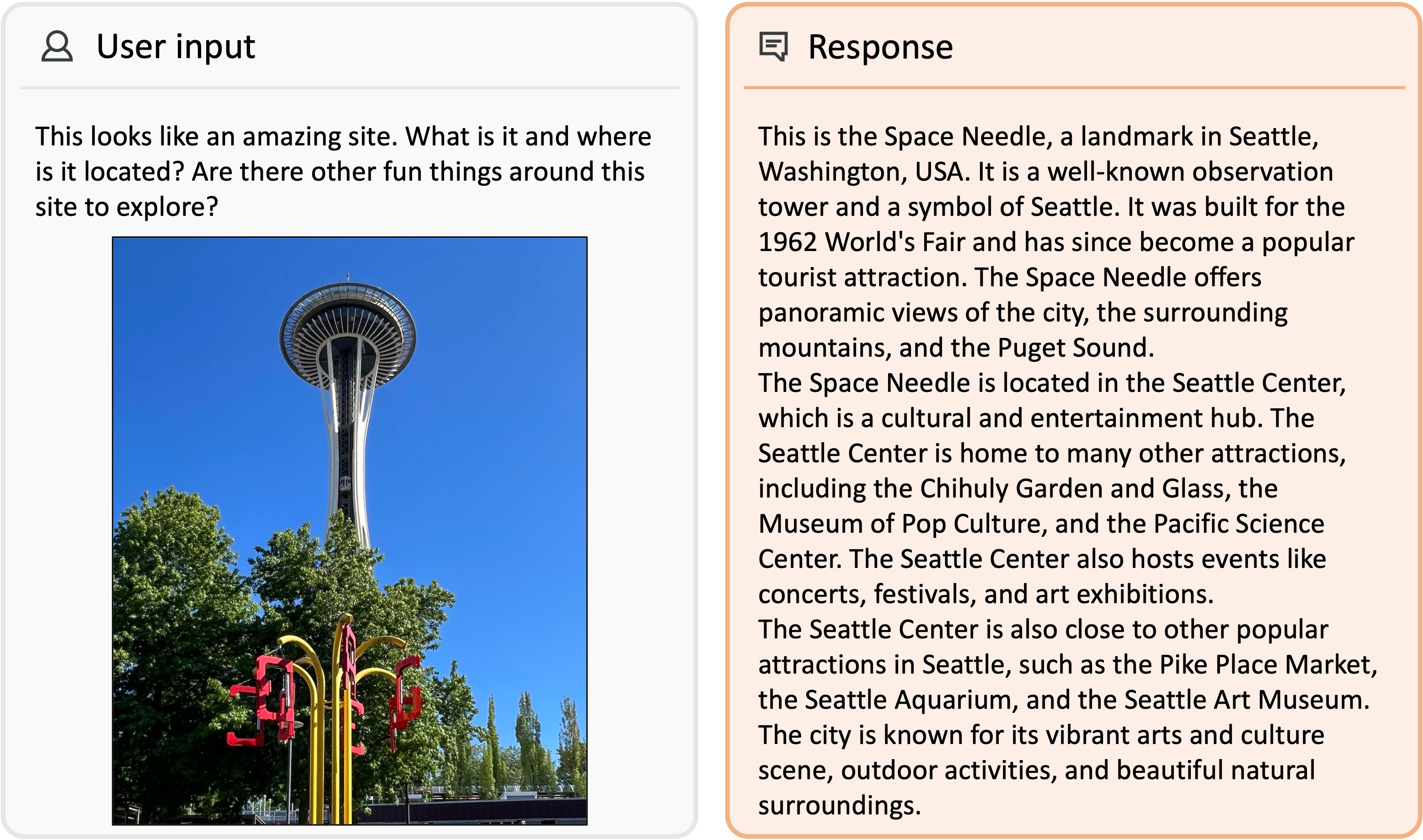}
    \caption{Created with Nova Pro. Photo taken by a team member.}
    \label{fig:qualitative_example_multimodal_intelligence}
\end{figure}

\begin{figure}\centering
    \includegraphics[width=1.0\linewidth]{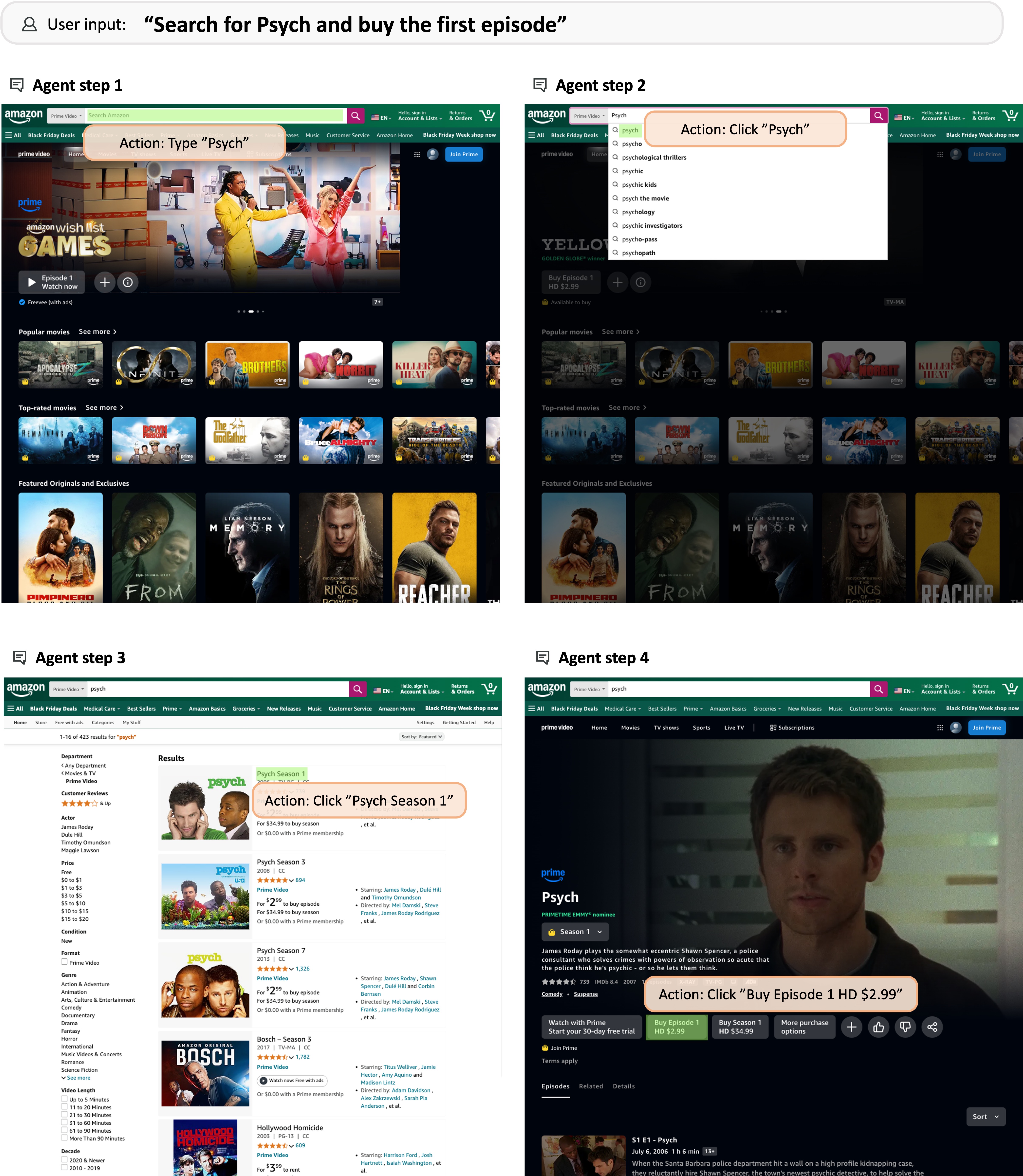}
    \caption{Example of a multimodal agent.}
    \label{fig:qualitative_example_multimodal_intelligence_appendix_agent}
\end{figure}

\begin{figure}\centering
    \includegraphics[width=1.0\linewidth]{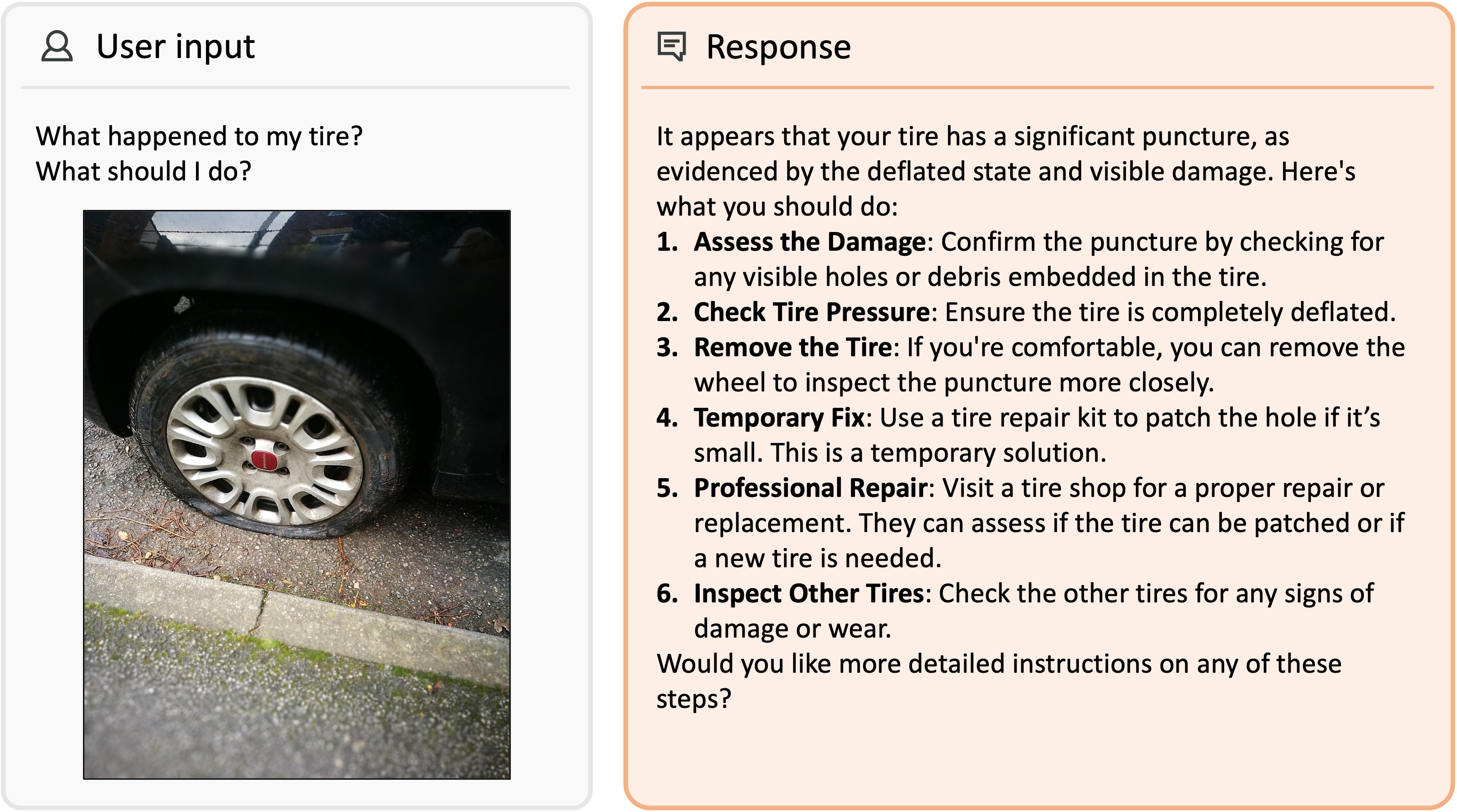}
    \caption{Created with Nova Pro. Image source: \cite{zhong2024law}.}
    \label{fig:qualitative_example_multimodal_intelligence_appendix_06}
\end{figure}

\begin{figure}\centering
    \includegraphics[width=1.0\linewidth]{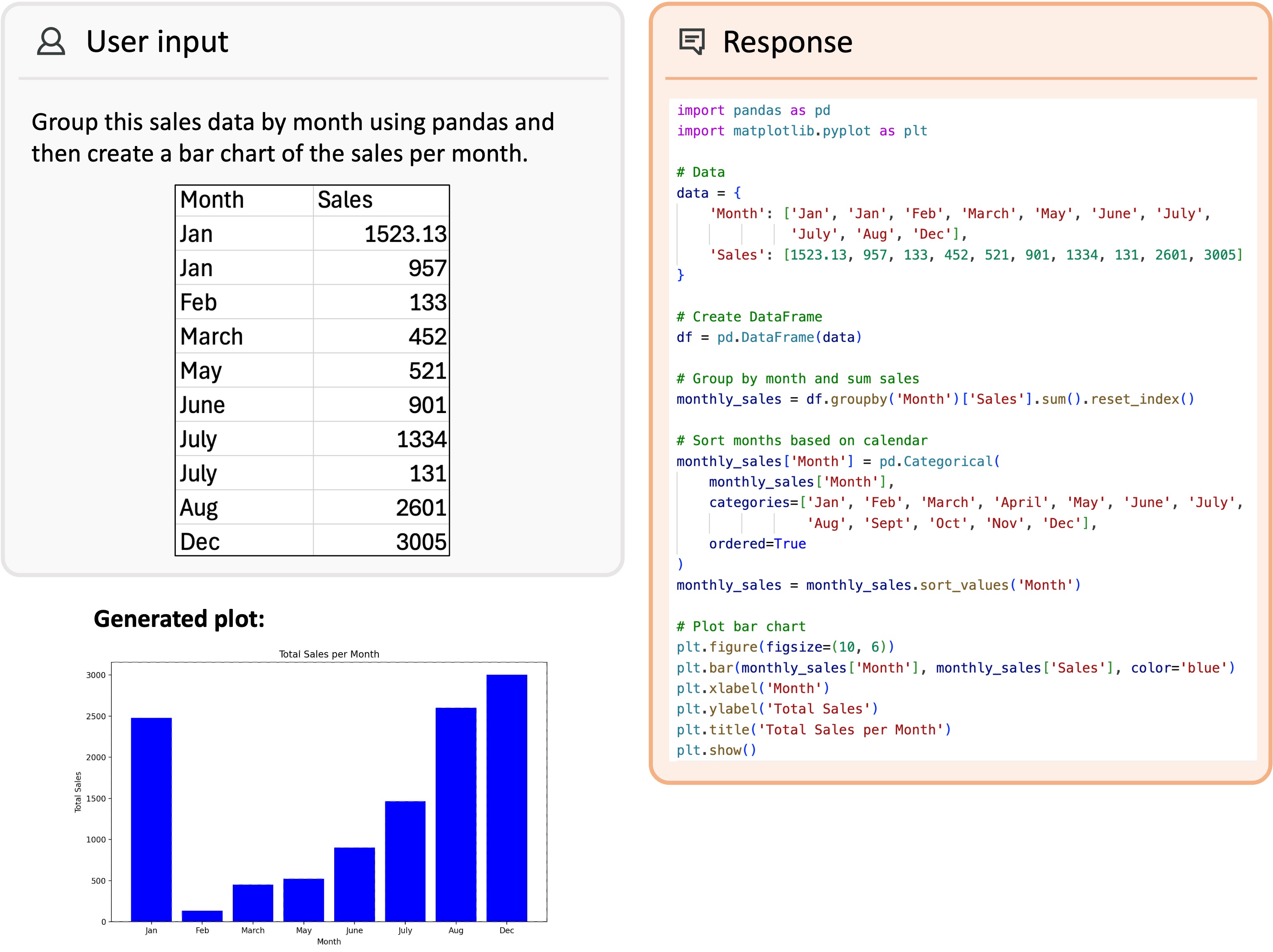}
    \caption{Created with Nova Pro.}
    \label{fig:qualitative_example_multimodal_intelligence_appendix_07}
\end{figure}

\begin{figure}\centering
    \includegraphics[width=1.0\linewidth]{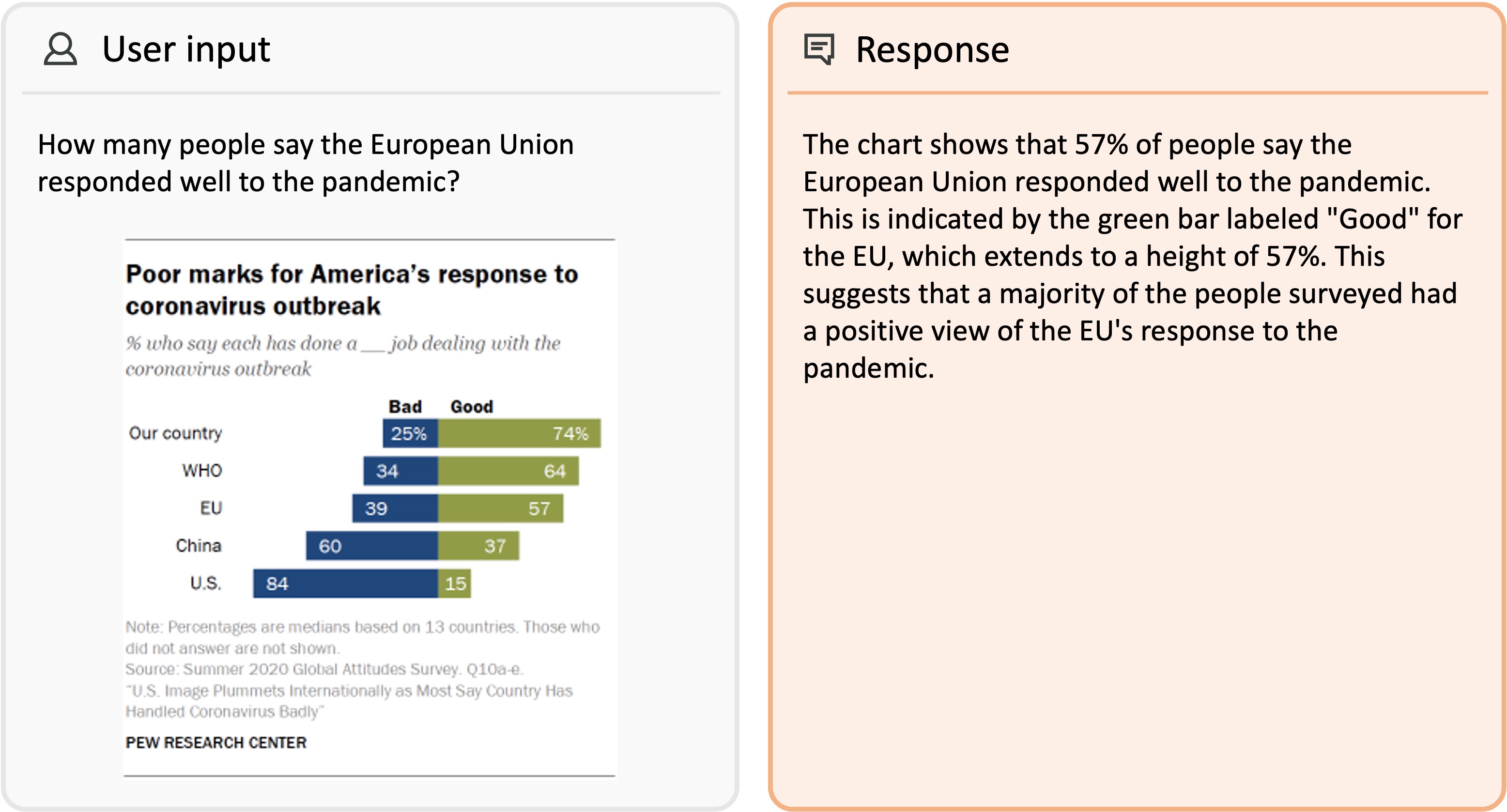}
    \caption{Created with Nova Lite. Image source: ChartQA dataset~\cite{chartqa}.}
    \label{fig:qualitative_example_multimodal_intelligence_appendix_05}
\end{figure}

\clearpage
\section{Correspondence and Contributors}

Please direct all correspondences to:

\begin{center}
\includegraphics[height=3ex]{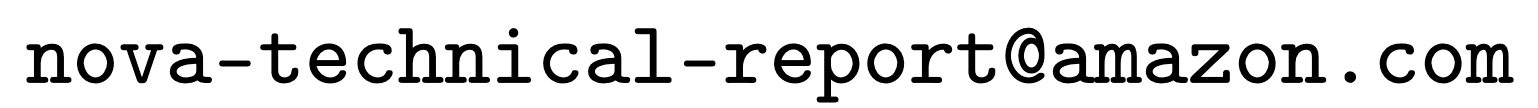}%
\end{center}

The Nova family of models were built by the Amazon Artificial General Intelligence (AGI) organization and partner teams.

When citing this report, please use ``Amazon AGI'' as the sole author, as shown in the bibtex entry below.

\begin{lstlisting}
@misc{novatechreport,
  author = {Amazon AGI},
  title  = {The Amazon Nova Family of Models: Technical Report and Model Card},
  year   = {2024},
  url    = {https://www.amazon.science/publications/the-amazon-nova-family-of-models-technical-report-and-model-card}
}
\end{lstlisting}

\subsection{Contributors}

The following individuals worked in the Nova program for at least one-fifth of its duration and measurably impacted one or more of the models or services described in this report.

\begin{multicols}{3}
\setlength{\parskip}{0pt}
Aaron Langford

Aayush Shah

Abhanshu Gupta

Abhimanyu Bhatter

Abhinav Goyal

Abhinav Mathur

Abhinav Mohanty

Abhishek Kumar

Abhishek Sethi

Abi Komma

Abner Pena

Achin Jain

Adam Kunysz

Adam Opyrchal

Adarsh Singh

Aditya Rawal

Adok Achar Budihal Prasad

Adrià de Gispert

Agnika Kumar

Aishwarya Aryamane

Ajay Nair

Akilan M

Akshaya Iyengar

Akshaya Vishnu Kudlu Shanbhogue

Alan He

Alessandra Cervone

Alex Loeb

Alex Zhang

Alexander Fu

Alexander Lisnichenko

Alexander Zhipa

Alexandros Potamianos

Ali Kebarighotbi

Aliakbar Daronkolaei

Alok Parmesh

Amanjot Kaur Samra

Ameen Khan

Amer Rez

Amir Saffari

Amit Agarwalla

Amit Jhindal

Amith Mamidala

Ammar Asmro

Amulya Ballakur

Anand Mishra

Anand Sridharan

Anastasiia Dubinina

Andre Lenz

Andreas Doerr

Andrew Keating

Andrew Leaver

Andrew Smith

Andrew Wirth

Andy Davey

Andy Rosenbaum

Andy Sohn

Angela Chan

Aniket Chakrabarti

Anil Ramakrishna

Anirban Roy

Anita Iyer

Anjali Narayan-Chen

Ankith Yennu

Anna Dabrowska

Anna Gawlowska

Anna Rumshisky

Anna Turek

Anoop Deoras

Anton Bezruchkin

Anup Prasad

Anupam Dewan

Anwith Kiran

Apoorv Gupta

Aram Galstyan

Aravind Manoharan

Arijit Biswas

Arindam Mandal

Arpit Gupta

Arsamkhan Pathan

Arun Nagarajan

Arushan Rajasekaram

Arvind Sundararajan

Ashwin Ganesan

Ashwin Swaminathan

Athanasios Mouchtaris

Audrey Champeau

Avik Ray

Ayush Jaiswal

Ayush Sharma

Bailey Keefer

Balamurugan Muthiah

Beatriz Leon-Millan

Ben Koopman

Ben Li

Benjamin Biggs

Benjamin Ott

Bhanu Vinzamuri

Bharath Venkatesh

Bhavana Ganesh

Bhoomit Vasani

Bill Byrne

Bill Hsu

Bincheng Wang

Blake King

Blazej Gorny

Bo Feng

Bo Zheng

Bodhisattwa Paul

Bofan Sun

Bofeng Luo

Bowen Chen

Bowen Xie

Boya Yu

Brendan Jugan

Brett Panosh

Brian Collins

Brian Thompson

Can Karakus

Can Liu

Carl Lambrecht

Carly Lin

Carolyn Wang

Carrie Yuan

Casey Loyda

Cezary Walczak

Chalapathi Choppa

Chandana Satya Prakash

Chankrisna Richy Meas

Charith Peris

Charles Recaido

Charlie Xu

Charul Sharma

Chase Kernan

Chayut Thanapirom

Chengwei Su

Chenhao Xu

Chenhao Yin

Chentao Ye

Chenyang Tao

Chethan Parameshwara

Ching-Yun Chang

Chong Li

Chris Hench

Chris Tran

Christophe Dupuy

Christopher Davis

Christopher DiPersio

Christos Christodoulopoulos

Christy Li

Chun Chen

Claudio Delli Bovi

Clement Chung

Cole Hawkins

Connor Harris

Corey Ropell

Cynthia He

DK Joo

Dae Yon Hwang

Dan Rosen

Daniel Elkind

Daniel Pressel

Daniel Zhang

Danielle Kimball

Daniil Sorokin

Dave Goodell

Davide Modolo

Dawei Zhu

Deepikaa Suresh

Deepti Ragha

Denis Filimonov

Denis Foo Kune

Denis Romasanta Rodriguez

Devamanyu Hazarika

Dhananjay Ram

Dhawal Parkar

Dhawal Patel

Dhwanil Desai

Dinesh Singh Rajput

Disha Sule

Diwakar Singh

Dmitriy Genzel

Dolly Goldenberg

Dongyi He

Dumitru Hanciu

Dushan Tharmal

Dzmitry Siankovich

Edi Cikovic

Edwin Abraham

Ekraam Sabir

Elliott Olson

Emmett Steven

Emre Barut

Eric Jackson

Ethan Wu

Evelyn Chen

Ezhilan Mahalingam

Fabian Triefenbach

Fan Yang

Fangyu Liu

Fanzi Wu

Faraz Tavakoli

Farhad Khozeimeh

Feiyang Niu

Felix Hieber

Feng Li

Firat Elbey

Florian Krebs

Florian Saupe

Florian Sprünken

Frank Fan

Furqan Khan

Gabriela De Vincenzo

Gagandeep Kang

George Ding

George He

George Yeung

Ghada Qaddoumi

Giannis Karamanolakis

Goeric Huybrechts

Gokul Maddali

Gonzalo Iglesias

Gordon McShane

Gozde Sahin

Guangtai Huang

Gukyeong Kwon

Gunnar A. Sigurdsson

Gurpreet Chadha

Gururaj Kosuru

Hagen Fuerstenau

Hah Hah

Haja Maideen

Hajime Hosokawa

Han Liu

Han-Kai Hsu

Hann Wang

Hao Li

Hao Yang

Haofeng Zhu

Haozheng Fan

Harman Singh

Harshavardhan Kaluvala

Hashim Saeed

He Xie

Helian Feng

Hendrix Luo

Hengzhi Pei

Henrik Nielsen

Hesam Ilati

Himanshu Patel

Hongshan Li

Hongzhou Lin

Hussain Raza

Ian Cullinan

Imre Kiss

Inbarasan Thangamani

Indrayani Fadnavis

Ionut Teodor Sorodoc

Irem Ertuerk

Iryna Yemialyanava

Ishan Soni

Ismail Jelal

Ivan Tse

Jack FitzGerald

Jack Zhao

Jackson Rothgeb

Jacky Lee

Jake Jung

Jakub Debski

Jakub Tomczak

James Jeun

James Sanders

Jason Crowley

Jay Lee

Jayakrishna Anvesh Paidy

Jayant Tiwari

Jean Farmer

Jeff Solinsky

Jenna Lau

Jeremy Savareese

Jerzy Zagorski

Ji Dai

Jiacheng (JC) Gu

Jiahui Li

Jian (Skyler) Zheng

Jianhua Lu

Jianhua Wang

Jiawei Dai

Jiawei Mo

Jiaxi Xu

Jie Liang

Jie Yang

Jim Logan

Jimit Majmudar

Jing Liu

Jinghong Miao

Jingru Yi

Jingyang Jin

Jiun-Yu Kao

Jixuan Wang

Jiyang Wang

Joe Pemberton

Joel Carlson

Joey Blundell

John Chin-Jew

John He

Jonathan Ho

Jonathan Hueser

Jonathan Lunt

Jooyoung Lee

Joshua Tan

Joyjit Chatterjee

Judith Gaspers

Jue Wang

Jun Fang

Jun Tang

Jun Wan

Jun Wu

Junlei Wang

Junyi Shi

Justin Chiu

Justin Satriano

Justin Yee

Jwala Dhamala

Jyoti Bansal

Kai Zhen

Kai-Wei Chang

Kaixiang Lin

Kalyan Raman

Kanthashree Mysore Sathyendra

Karabo Moroe

Karan Bhandarkar

Karan Kothari

Karolina Owczarzak

Karthick Gopalswamy

Karthick Ravi

Karthik Ramakrishnan

Karthika Arumugam

Kartik Mehta

Katarzyna Konczalska

Kavya Ravikumar

Ke Tran

Kechen Qin

Kelin Li

Kelvin Li

Ketan Kulkarni

Kevin Angelo Rodrigues

Keyur Patel

Khadige Abboud

Kiana Hajebi

Klaus Reiter

Kris Schultz

Krishna Anisetty

Krishna Kotnana

Kristen Li

Kruthi Channamallikarjuna

Krzysztof Jakubczyk

Kuba Pierewoj

Kunal Pal

Kunwar Srivastav

Kyle Bannerman

Lahari Poddar

Lakshmi Prasad

Larry Tseng

Laxmikant Naik

Leena Chennuru Vankadara

Lenon Minorics

Leo Liu

Leonard Lausen

Leonardo F. R. Ribeiro

Li Zhang

Lili Gehorsam

Ling Qi

Lisa Bauer

Lori Knapp

Lu Zeng

Lucas Tong

Lulu Wong

Luoxin Chen

Maciej Rudnicki

Mahdi Namazifar

Mahesh Jaliminche

Maira Ladeira Tanke

Manasi Gupta

Mandeep Ahlawat

Mani Khanuja

Mani Sundaram

Marcin Leyk

Mariusz Momotko

Markus Boese

Markus Dreyer

Markus Mueller

Mason Fu

Mateusz Górski

Mateusz Mastalerczyk

Matias Mora

Matt Johnson

Matt Scott

Matthew Wen

Max Barysau

Maya Boumerdassi

Maya Krishnan

Mayank Gupta

Mayank Hirani

Mayank Kulkarni

Meganathan Narayanasamy

Melanie Bradford

Melanie Gens

Melissa Burke

Meng Jin

Miao Chen

Michael Denkowski

Michael Heymel

Michael Krestyaninov

Michal Obirek

Michalina Wichorowska

Michał Miotk

Milosz Watroba

Mingyi Hong

Mingzhi Yu

Miranda Liu

Mohamed Gouda

Mohammad El-Shabani

Mohammad Ghavamzadeh

Mohit Bansal

Morteza Ziyadi

Nan Xia

Nathan Susanj

Nav Bhasin

Neha Goswami

Nehal Belgamwar

Nicolas Anastassacos

Nicolas Bergeron

Nidhi Jain

Nihal Jain

Niharika Chopparapu

Nik Xu

Nikko Strom

Nikolaos Malandrakis

Nimisha Mishra

Ninad Parkhi

Ninareh Mehrabi

Nishita Sant

Nishtha Gupta

Nitesh Sekhar

Nithin Rajeev

Nithish Raja Chidambaram

Nitish Dhar

Noor Bhagwagar

Noy Konforty

Omar Babu

Omid Razavi

Orchid Majumder

Osama Dar

Oscar Hsu

Pablo Kvitca

Pallavi Pandey

Parker Seegmiller

Patrick Lange

Paul Ferraro

Payal Motwani

Pegah Kharazmi

Pei Wang

Pengfei Liu

Peter Bradtke

Peter Götz

Peter Zhou

Pichao Wang

Piotr Poskart

Pooja Sonawane

Pradeep Natarajan

Pradyun Ramadorai

Pralam Shah

Prasad Nirantar

Prasanthi Chavali

Prashan Wanigasekara

Prashant Saraf

Prashun Dey

Pratyush Pant

Prerak Pradhan

Preyaa Patel

Priyanka Dadlani

Prudhvee Narasimha Sadha

Qi Dong

Qian Hu

Qiaozi (QZ) Gao

Qing Liu

Quinn Lam

Quynh Do

R. Manmatha

Rachel Willis

Rafael Liu

Rafal Ellert

Rafal Kalinski

Rafi Al Attrach

Ragha Prasad

Ragini Prasad

Raguvir Kunani

Rahul Gupta

Rahul Sharma

Rahul Tewari

Rajaganesh Baskaran

Rajan Singh

Rajiv Gupta

Rajiv Reddy

Rajshekhar Das

Rakesh Chada

Rakesh Vaideeswaran Mahesh

Ram Chandrasekaran

Ramesh Nallapati

Ran Xue

Rashmi Gangadharaiah

Ravi Rachakonda

Renxian Zhang

Rexhina Blloshmi

Rishabh Agrawal

Robert Enyedi

Robert Lowe

Robik Shrestha

Robinson Piramuthu

Rohail Asad

Rohan Khanna

Rohan Mukherjee

Rohit Mittal

Rohit Prasad

Rohith Mysore Vijaya Kumar

Ron Diamant

Ruchita Gupta

Ruiwen Li

Ruoying Li

Rushabh Fegade

Ruxu Zhang

Ryan Arbow

Ryan Chen

Ryan Gabbard

Ryan Hoium

Ryan King

Sabarishkumar Iyer

Sachal Malick

Sahar Movaghati

Sai Balakavi

Sai Jakka

Sai Kashyap Paruvelli

Sai Muralidhar Jayanthi

Saicharan Shriram Mujumdar

Sainyam Kapoor

Sajjad Beygi

Saket Dingliwal

Saleh Soltan

Sam Ricklin

Sam Tucker

Sameer Sinha

Samridhi Choudhary

Samson Tan

Samuel Broscheit

Samuel Schulter

Sanchit Agarwal

Sandeep Atluri

Sander Valstar

Sanjana Shankar

Sanyukta Sanyukta

Sarthak Khanna

Sarvpriye Khetrapal

Satish Janakiraman

Saumil Shah

Saurabh Akolkar

Saurabh Giri

Saurabh Khandelwal

Saurabh Pawar

Saurabh Sahu

Sean Huang

Sejun Ra

Senthilkumar Gopal

Sergei Dobroshinsky

Shadi Saba

Shamik Roy

Shamit Lal

Shankar Ananthakrishnan

Sharon Li

Shashwat Srijan

Shekhar Bhide

Sheng Long Tang

Sheng Zha

Shereen Oraby

Sherif Mostafa

Shiqi Li

Shishir Bharathi

Shivam Prakash

Shiyuan Huang

Shreya Yembarwar

Shreyas Pansare

Shreyas Subramanian

Shrijeet Joshi

Shuai Liu

Shuai Tang

Shubham Chandak

Shubham Garg

Shubham Katiyar

Shubham Mehta

Shubham Srivastav

Shuo Yang

Siddalingesha D S

Siddharth Choudhary

Siddharth Singh Senger

Simon Babb

Sina Moeini

Siqi Deng

Siva Loganathan

Slawomir Domagala

Sneha Narkar

Sneha Wadhwa

Songyang Zhang

Songyao Jiang

Sony Trenous

Soumajyoti Sarkar

Soumya Saha

Sourabh Reddy

Sourav Dokania

Spurthideepika Sandiri

Spyros Matsoukas

Sravan Bodapati

Sri Harsha Reddy Wdaru

Sridevi Yagati Venkateshdatta

Srikanth Ronanki

Srinivasan R Veeravanallur

Sriram Venkatapathy

Sriramprabhu Sankaraguru

Sruthi Gorantla

Sruthi Karuturi

Stefan Schroedl

Subendhu Rongali

Subhasis Kundu

Suhaila Shakiah

Sukriti Tiwari

Sumit Bharti

Sumita Sami

Sumith Mathew

Sunny Yu

Sunwoo Kim

Suraj Bajirao Malode

Susana Cumplido Riel

Swapnil Palod

Swastik Roy

Syed Furqhan

Tagyoung Chung

Takuma Yoshitani

Taojiannan Yang

Tejaswi Chillakura

Tejwant Bajwa

Temi Lajumoke

Thanh Tran

Thomas Gueudre

Thomas Jung

Tianhui Li

Tim Seemman

Timothy Leffel

Tingting Xiang

Tirth Patel

Tobias Domhan

Tobias Falke

Toby Guo

Tom Li

Tomasz Horszczaruk

Tomasz Jedynak

Tushar Kulkarni

Tyst Marin

Tytus Metrycki

Tzu-Yen Wang

Umang Jain

Upendra Singh

Utkarsh Chirimar

Vaibhav Gupta

Vanshil Shah

Varad Deshpande

Varad Gunjal

Varsha Srikeshava

Varsha Vivek

Varun Bharadwaj

Varun Gangal

Varun Kumar

Venkatesh Elango

Vicente Ordonez

Victor Soto

Vignesh Radhakrishnan

Vihang Patel

Vikram Singh

Vinay Varma Kolanuvada

Vinayshekhar Bannihatti Kumar

Vincent Auvray

Vincent Cartillier

Vincent Ponzo

Violet Peng

Vishal Khandelwal

Vishal Naik

Vishvesh Sahasrabudhe

Vitaliy Korolev

Vivek Gokuladas

Vivek Madan

Vivek Subramanian

Volkan Cevher

Vrinda Gupta

Wael Hamza

Wei Zhang

Weitong Ruan

Weiwei Cheng

Wen Zhang

Wenbo Zhao

Wenyan Yao

Wenzhuo Ouyang

Wesley Dashner

William Campbell

William Lin

Willian Martin

Wyatt Pearson

Xiang Jiang

Xiangxing Lu

Xiangyang Shi

Xianwen Peng

Xiaofeng Gao

Xiaoge Jiang

Xiaohan Fei

Xiaohui Wang

Xiaozhou Joey Zhou

Xin Feng

Xinyan Zhao

Xinyao Wang

Xinyu Li

Xu Zhang

Xuan Wang

Xuandi Fu

Xueling Yuan

Xuning Wang

Yadunandana Rao

Yair Tavizon

Yan Rossiytsev

Yanbei Chen

Yang Liu

Yang Zou

Yangsook Park

Yannick Versley

Yanyan Zhang

Yash Patel

Yen-Cheng Lu

Yi Pan

Yi-Hsiang (Sean) Lai

Yichen Hu

Yida Wang

Yiheng Zhou

Yilin Xiang

Ying Shi

Ying Wang

Yishai Galatzer

Yongxin Wang

Yorick Shen

Yuchen Sun

Yudi Purwatama

Yue (Rex) Wu

Yue Gu

Yuechun Wang

Yujun Zeng

Yuncong Chen

Yunke Zhou

Yusheng Xie

Yvon Guy

Zbigniew Ambrozinski

Zhaowei Cai

Zhen Zhang

Zheng Wang

Zhenghui Jin

Zhewei Zhao

Zhiheng Li

Zhiheng Luo

Zhikang Zhang

Zhilin Fang

Zhiqi Bu

Zhiyuan Wang

Zhizhong Li

Zijian Wang

Zimeng (Chris) Qiu

Zishi Li

\end{multicols}

\subsection{Acknowledgements}

We would like to acknowledge the following individuals who supported the development of the Nova models and services during the Nova program.

\begin{multicols}{3}
\setlength{\parskip}{0pt}
Abdelrahman Badawy

Abtin Rasoulian

Adam Baranowski

Aishwarya Kore

Aishwarya Padmakumar

Alain Krok

Alex Mould

Alex Sun

Alexandros Papangelis

Alfred Shen

Amaran Asokkumar

Amiya Chakraborty

Anastasios Alexandridis

Angeliki Metallinou

Anila Joshi

Anup Katariya

Arda Keskiner

Avinash Venkatagiri

Aya Elzoheiry

Baishali Chaudhury

Ben Friebe

Bigad Soleiman

Bob Li

Brad Porter

Brian Chou

Brian Yost

Burak Gozluklu

Chad Connally

Chris Azer

Chris Beauchene

Chris Greenwood

Chris Johnson

Clay Cheng

Craig Rowland

Di Jin

Di Wu

Diego Socolinsky

Don Kretsch

Dylan Martin

Emma Lister

Eva Lasarcyk

Evan Kravitz

Federico D'Alessio

Flora Wang

Francisco Calderon Rodriguez

Gamaleldin Elsayed

Gaurav Rele

Gaurav Sukhatme

Gourav Datta

Hadrien Glaude

Hanbo Wang

Hans Hoeijmaker

Haotian An

Harpreet Cheema

Harshit Pande

Hongbin Zheng

Huda Khayrallah

Isaac Privitera

Jacob Zhiyuan Fang

Jady Liu

Jae Oh Woo

Jamal Saboune

James Park

Jianbo Yuan

Jianwei Feng

Jie Li

Jinwoo Park

Johan Esbjorner

Jonathan makunga

JoonHyung Kim

Jorge Beltran

Jose Garrido Ramas

Julie Baca

Justin Lewis

Kamran Razi

Kangyan Liu

Kasana Mahesh

Kelvin Qian

Kyle Goehner

Kyle Saggar

Laith Al-Saadoon

Lei Sun

Lily Liao

Long Chen

Lukacs Ablonczy

Luke Luneau

Maciej Eichler

Mallory McManamon

Manju Arakere

Matt McCoy

Matthew Chang

Meghal Varia

Meghana Ashok

Melanie Li

Mifu Suzuki

Negin Sokhandan

Nick Biso

Nico Bishop

Nicolle Borges

Palash Goyal

Parker Coleman

Paul Sumarokov

Pavel Kveton

Philipp Lerche

Pratibha Kumari

Rahul Agarwal

Rahul Ghosh

Rahul Kulkarni

Raj Kumar

Ramana Keerthi

Rams Sundaram

Raymond Fang

Reethika Kesani

Ryan Razkenari

Sarath Krishnan

Scott Patten

Seokhwan Kim

Sepehr Eghbali

Sergey Pugachev

Sertan Alkan

Shailav Taneja

Sheamus Punch

Shikib Mehri

Shilpa Singh

Shraddha Ravishankar

Sijia Liu

Sitanshu Gupta

Sol Vesdapunt

Spencer Romo

Sravya Uppu

Srivani Kambhampati

Stephanie Xie

Sujitha Martin

Sungjin Lee

Sungmin Hong

Tanner McRae

Thomas Patterson

Tina Li

Tom Liang

Trong Nguyen

Vasudev Mahesh Purandare

Vidya Sagar Ravipati

Vu San Ha Huynh

Weijuan Wu

Xiaolong Li

Xinyi Xu

Yaroslav Nechaev

Yuan Tian

Yunfei Bai

Zach Hille

Ziyan Tian

\end{multicols}

\end{document}